\documentclass[twocolumn]{article}

\pdfoutput=1

\usepackage[left=0.71in,top=0.95in,right=0.71in,bottom=0.75in]{geometry}
\usepackage{caption}
\usepackage{epsfig}
\usepackage{subfigure}
\usepackage{amsmath}

\newcommand{\etal}{et al.\@}
\newcommand{\ie}{i.e.\@ }

\begin{document}

\title{Color-opponent mechanisms for local hue encoding in a hierarchical framework}

\author{Paria Mehrani\thanks{The two authors contributed equally to this work.}, Andrei Mouraviev\footnotemark[1], Oscar J. Avella Gonzalez, John K. Tsotsos\\
	paria@cse.yorku.ca, andrei.mouraviev@gmail.com, \\ ojavellag@cse.yorku.ca , tsotsos@cse.yorku.ca }

\date{}

\maketitle

\providecommand{\keywords}[1]{\textbf{\textit{Keywords---}} #1}

\begin{abstract}
A biologically plausible computational model for color representation is introduced. We present a mechanistic hierarchical model of neurons that not only successfully encodes local hue, but also explicitly reveals how the contributions of each visual cortical layer participating in the process can lead to a hue representation. Our proposed model benefits from studies on the visual cortex and builds a network of single-opponent and hue-selective neurons. Local hue encoding is achieved through gradually increasing nonlinearity in terms of cone inputs to single-opponent cells. We demonstrate that our model's single-opponent neurons have wide tuning curves, while the hue-selective neurons in our model V4 layer exhibit narrower tunings, resembling those in V4 of the primate visual system. Our simulation experiments suggest that neurons in V4 or later layers have the capacity of encoding unique hues. Moreover, with a few examples, we present the possibility of spanning the infinite space of physical hues by combining the hue-selective neurons in our model.
\end{abstract}


\section{Introduction}
The goal of this study is to introduce a biologically-plausible computational color processing model that as Brown~\cite{Brown2014tale} argues, helps in ``understand[ing] how the elements of the brain work together to form functional units and ultimately generate the complex cognitive behaviors we study''. For this purpose, we propose a hierarchical model of neurons, each layer of which corresponds to one processing layer in the brain. We show that the behavior of our model neurons matches that of the corresponding layers in the brain. Our hierarchical framework suggests a computational mechanism for color representation in the ventral stream.

\begin{figure}
	\centering
	\includegraphics[width = 0.35\textwidth]{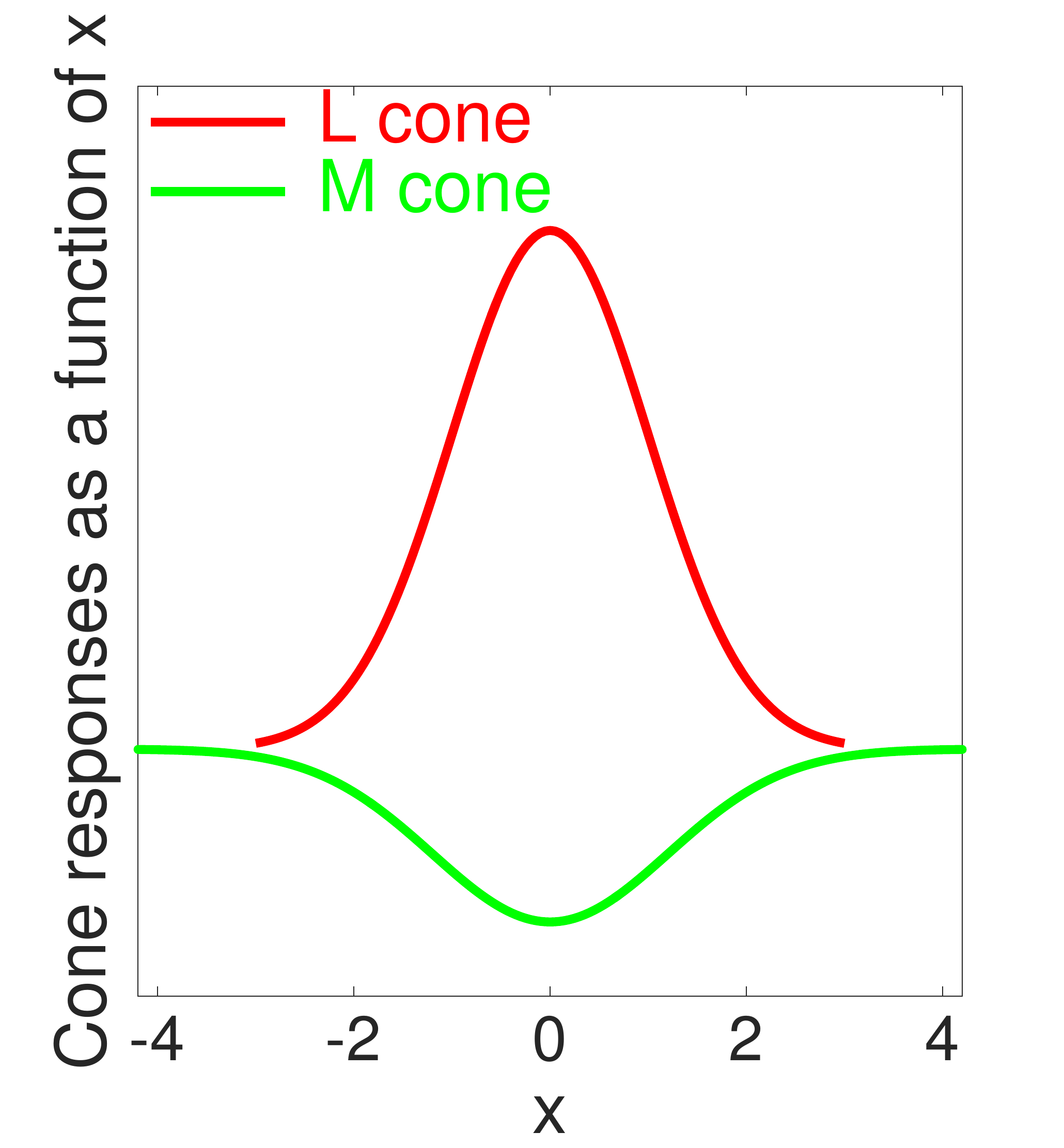}
	\caption{Spatial profile of a single-opponent $\text{L+M-}$ cell. The receptive field of this cells receives positive contributions from L cones and negative ones from M cones. The spatial extent of these cone cells are different and determines the mechanism for this cell. Figure adapted from~\protect\cite{shapley2002neural}.}
	\label{fig:single_opp}
\end{figure}

Studies on the human visual system confirm that color encoding starts in a three-dimensional space with three types of cone cells, each sensitive to a certain band of wavelengths. These wavelengths, categorized according to the cone sensitivities to short \textbf{(S)}, medium \textbf{(M)} and long \textbf{(L)} wavelengths, form the LMS color space.
Color encoding in the visual system gets more complicated in higher layers. Cones send feedforward signals to LGN, whose cells are characterized by their opponent inputs from different cones~\cite{Reid2002}. Reid and Shapley demonstrated that LGN cells with single-opponent receptive fields receive opponent cone inputs that overlap in the visual field with different extents. An example of such a spatial profile is illustrated in Figure~\ref{fig:single_opp}. They also mapped the contributions of cones to LGN receptive fields. 
Color opponent mechanisms were the basis of the ``Opponent Process Theory'' of Hering~\cite{Hering}, in which he also introduced unique hues, those pure colors unmixed with other colors. In his theory, there are four unique hues: red, green, yellow and blue that were believed to be encoded by cone-opponent processes. Later studies~\cite{Derrington1984chromatic,DeValois2000physio,Sun2006ConeInput}, however, confirmed that the cone-opponent mechanisms of earlier processing stages do not correspond to Hering's red vs. green and yellow vs. blue opponent processes. In fact, they observed that the color coding in these regions is organized along the two dimensions of the MacLeod and Boyton~\cite{Macleod_Boyton} diagram. That is, along L vs M and S vs LM axes. In a study on color appearance in normal vision subjects, Webster \etal~\cite{Webster2000variations} found that while unique red is close to the +L axis, unique green, yellow and blue hues cluster around intermediate directions in the MacLeod-Boyton diagram. A similar study~\cite{Wuerger2005} suggested that the encoding of unique hues, unlike the tuning of LGN neurons, needs higher order mechanisms such as a piecewise linear model in terms of cone inputs. 

Even though most studies agree LGN tunings do not encode unique hues, there exists no such unanimous agreement about V1 and V2 tunings. Lennie \etal~\cite{Lennie1990chromatic} observed similar chromatic selectivity as LGN neurons in V1 cells and suggested a rectified sum of the three cone types describing the responses of V1 neurons. Hanazawa \etal \cite{Hanazawa} found both linear and nonlinear V1 and V2 responses with respect to the three cone types, with gradual increase in percentage of nonlinearity from LGN to higher layers. A similar finding by Kuriki \etal~\cite{Kuriki2015fMRIhue} suggested that there are selectivities to both intermediate and cone-opponent hues in areas V1, V2, V3 and V4, with more diverse selectivity in V4 compared to the other areas. De Valois \etal\cite{DeValois2000physio} suggested perceptual hues can be encoded in V1, obtained by combining LGN responses in a nonlinear fashion. Wachtler~\etal \cite{Wachtler2003} found that the tunings of V1 neurons are different from LGN, and that the responses in V1 are affected by context as well as remote color patches in both suppressive and enhancement manners.

In a set of studies~\cite{johnson2004cone,shapley2002neural}, three different types of cells in V1 were identified: luminance, color-luminance, and color-preferring neurons. Johnson~\etal~\cite{johnson2004cone} suggested that cells with a single-opponent receptive field encode local hue, while double-opponent cells signal color contrast.

Color encoding in higher layers is less understood. Millimeter-sized islands in extrastriate cortex, called globs, were shown to have cells with luminance-invariant color tunings in the Macaque extrastriate cortex~\cite{Conway2007}. Additionally, cells in different globs represent distinct visual field locations. Conway and Tsao~\cite{Conway_Tsao2009} observed that neurons in a single glob have various color preferences and those in adjacent globs showed similar color tunings. Consequently, they suggested that cells in a glob are clustered by color preference and form the hypothesized color columns of Barlow~\cite{Barlow1986}. In contrast with previous findings, Namima~\etal~\cite{Namima2014} found that not only the responses of the neurons were luminance-dependent, but also that the effect of luminance in responses of neurons in V4, AIT and PIT varies from one stimulus color to another.

\begin{figure}
	\centering
	\centering
	\subfigure[Combined map of hue patches in three clusters.]{\includegraphics[width=0.2\textwidth]{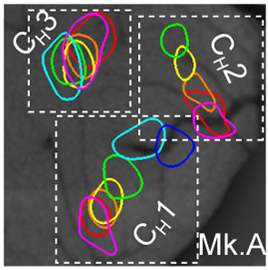}\label{subFig:colorMap_clusters}}
	\subfigure[Larger view of a cluster.]{\includegraphics[width=0.2\textwidth]{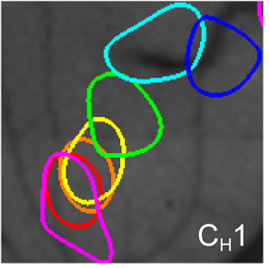}\label{subFig:colorMap_cluster1}}
	\caption{Color map of V4 neurons in three clusters of patches (adapted from~\protect\cite{li2014Map}).}
	\label{fig:colorMap}
\end{figure}

A recent study~\cite{li2014Map} on V4 examined how perceptual colors are mapped in this layer. They identified clusters of hue-selective patches, which they called ``rainbows of patches''. An example of an identified map of three clusters is shown in Figure~\ref{subFig:colorMap_clusters}. In Figure~\ref{subFig:colorMap_cluster1} a larger view of one of the clusters is shown. 
A close investigation on these patches revealed that their sequential representation follows the color order in the HSL space, hence a ``rainbow'' of patches is observed. In addition, they observed that each color activated 1-4 overlapping patches and that neighboring patches are activated for similar hues.  Moreover, they found that patches in each cluster have the same local visual field and that same-hue patches across clusters have a great overlap in their local visual fields. Following these observations, they concluded that any hue in the visual field is represented by 1-4 patches, and different hues are represented by different multi-patch patterns. Comparable findings in V2 were reported in~\cite{XiaoTopoV2}. They also propose a distributed and combinatorial color representation as the solution for encoding the large space of physical colors, given the limited number of neurons in each cortical color map. Representation of the color space in globs was also studied by Bohon~\etal~\cite{Bohon2016} who found that glob populations result in a uniform representation of the color space with bias toward ``warm'' colors. They also observed these that cells have nonlinear narrow tunings. Similarly, Schein and Desimnone~\cite{Schein1990_silent_surround} and Zeki~\cite{Zeki1980} remarked on narrow-band tunings in V4 cells.

Transformation mechanisms from cone-opponent responses to unique hues is still unclear. From responses of IT neurons, Zaidi~\etal~\cite{Zaidi2014} could accurately decode colors without any reference to unique hues. They also observed no significance of unique hues in human subjects. However, Stoughton and Conway~\cite{Stoughton2008}, similar to Zeki~\cite{Zeki1980} and Komatsu~\etal\cite{Komatsu1992}, observed that even though neurons in PIT show selectivities to all hue angles in color space, there are more neurons selective to those close to unique hues. Hence, they suggest an encoding for unique hues in higher layers, and rejected such representations in V1 and V2. The choice of stimuli for recordings in~\cite{Stoughton2008} was challenged by Mollon~\cite{Mollon2009} commenting that it is still unclear whether or not unique hues are represented in IT.

Among all the attempts to understand the neural processes for transformation from cone-opponency to perceptual colors, a number of computational models tried to suggest mechanisms for this problem and other aspects of color representation in higher areas. For example, the role of context in color constancy and color induction in V4 was studied by Dufort and Lumsden~\cite{Dufort_Lumsden}. Their model combines L+M- and S+(L+M)- LGN neurons to achieve color constancy. Later, Courtney~\etal~\cite{Courtney1995} achieved this goal in V4 by means of a ``silent'' surround~\cite{Schein1990_silent_surround} for providing context, while \cite{Wray1996} introduced a model of feedforward, feedback and lateral connections for this purpose.

Among the existing computational models, the most relevant model to ours would be the work of \cite{DeValois1993multi}, in which they introduced a network consisting of three stages: cone stage, cone-opponent stage, and perceptual color representation stage. Introduction of the third stage was inspired by the noted discrepancy between the axes of cone-opponency and perceptual opponency mentioned earlier. They proposed that S-modulations in the third stage rotate the cone-opponent axes in a way that results in perceptual-opponent axes. The third stage responses are obtained by a linear combination of cone-opponent signals, followed by a half-wave rectification. De Valois~\etal did not explicitly mention to which of the visual processing layers in the brain this third stage corresponds. Here, we examine two possibilities. For the first possibility, suppose the perceptual hue stage corresponds to V1, which is downstream of LGN. This is not an invalid assumption as this stage combines the linearly modeled cone-opponent responses. In this case, perceptual hues are encoded in V1, as found in a later study by the authors~\cite{DeValois2000physio}, in contrast with other findings for this visual area~\cite{Conway2001spatial,Conway2006spatial,Kiper1997V2,Lennie1990chromatic}. For the second possibility, suppose their third stage is modeling the neurons in V4. In that case, they provide no explicit model of V1 or V2 processes, even though these levels are believed to play the role of gradually increasing the nonlineraity in the processing of cone inputs~\cite{Hanazawa} in terms of cone inputs. Omitting the nonlinear layers of V1 and V2 results in wide tunings, an outcome which is not supported by previous and recent discoveries~\cite{Bohon2016,Zeki1980}. In short, we would like to emphasize that the De Valois \& De Valois model is a one-layer formalism of perceptual hue encoding, or in other words, the totality of processing is compressed into a single layer process. The end result may indeed provide a suitable model in the sense of its input-output characterization. However, it does not make an explicit statement about what each of the processing layers of the visual cortex are contributing to the overall result. Our goal differs in that we seek to build a mechanistic model that not only represents the local hue, but also computes this encoding, making explicit how each visual cortical layer participates in the process.


In this paper, we focus on modeling hue. 
Our model, inspired by neural mechanisms in the visual system, develops the representation in a hierarchical framework. This model, implemented upto and including layer V4, encodes local hues and shows activities, in V4, comparable to those of~\cite{li2014Map}. 
Our proposed network differs from that of~\cite{DeValois1993multi} as it explicitly models neurons in each of LGN, V1, V2, and V4 areas. Additionally, nonlinearity is gradually increased from one layer to another as observed by~\cite{Hanazawa}. While model neurons in V1 and V2 have similar selectivities to those of LGN~\cite{Lennie1990chromatic}, our V4 neurons exhibit narrower tunings~\cite{Bohon2016,Zeki1980}.
In order to verify the validity of our hue-selective neurons, we carried out a set of simulation experiments, which demonstrate that our model hue-selective neurons have behaviour similar to those in the visual area V4 and that their representation pattern mimics that of V4 neurons. We also suggest, through a few examples, a mechanism for encoding the vast space of physical colors by combining the pattern of activities of these hue-selective neurons.

In the following section we describe the details of each layer in our model. Next, we demonstrate results of our model in three different experiments, followed by a discussion.

\section{Method}
\begin{figure}
	\centering
	\includegraphics[width = 0.45\textwidth]{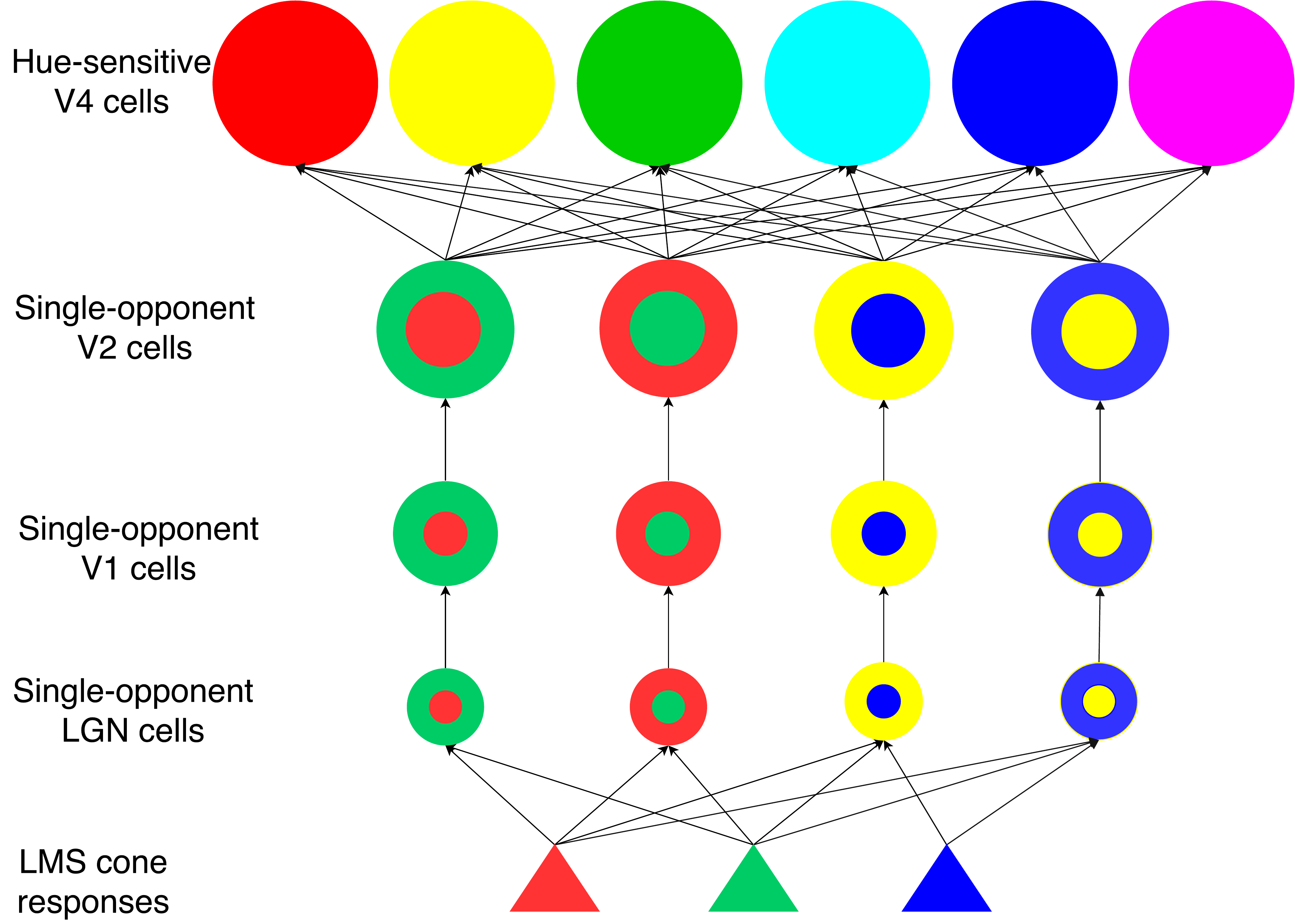}
	\caption{An illustration of the proposed hierarchical model for local hue representation. The first layer shows the input to the network with the LMS cone activations. A combination of cone responses with opposite signs result in neurons with single-opponent receptive fields in model layers LGN, V1, and V2. Note that the receptive field sizes increase, figuratively here, from one layer to the next as described in the text. The top layer of the model consists of hue-selective V4 neurons (Best seen in color).}
	\label{fig:network}
\end{figure}

Figure~\ref{fig:network} demonstrates the proposed model for color representation. 
In what follows, the input to the model will be shown in the RGB color space, rather than LMS, for ease of interpretation by the reader. In the event that the presented stimulus was available in RGB, we first performed a conversion into LMS channels using the transformation algorithm proposed by ~\cite{LMS_conversion} (we used the C code provided by the authors). As a result, one can think of the presented stimulus to the network as the activations of three cone types. 

Our model was implemented in TarzaNN~\cite{Rothenstein2005}. The neurons in all layers are linearly rectified. The rectification was performed using:
\begin{equation}
\phi(P) = 
	\begin{cases}
		\tau, & \text{if}\ mP+b < \tau\\
		mP+b , & \text{if}\ \tau \leq mP+b \leq s\\
		1, & \text{otherwise}, 
	\end{cases}
	\label{eq:rectifier}
\end{equation}
where $P$ is neuron activity, and $m$ and $b$ are the slope and base spike rate respectively, $\tau$ is a lower threshold of activities and $s$ represents the saturation threshold. This rectifier maps responses to $[\tau, 1]$. Depending on the settings of parameters $\tau$ and $s$, and the range of activations for the model neurons, the rectifier might vary from being linear to nonlinear. Wherever this rectifier is employed in the rest of the paper, we mention the settings of the parameters, and that whether the rectifier caused function of the neuron activations to become linear or nonlinear.

The input to the hierarchical network was always resized to $256\times256$ pixels. The receptive field sizes, following~\protect\cite{freeman2011metamers}, double from one layer to the one above. Specifically, the receptive field sizes we employed were $19\times19$, $38\times38$, $76\times76$, and $152\times152$ pixels for LGN, V1, V2, and V4 layers respectively.

In each layer of the network shown in Figure~\ref{fig:network}, one single cell of each type is shown in order to illustrate how the hue-selective neurons are modeled in the hierarchy. 
\begin{figure}[t!]
	\centering
	\includegraphics[width = 0.45\textwidth]{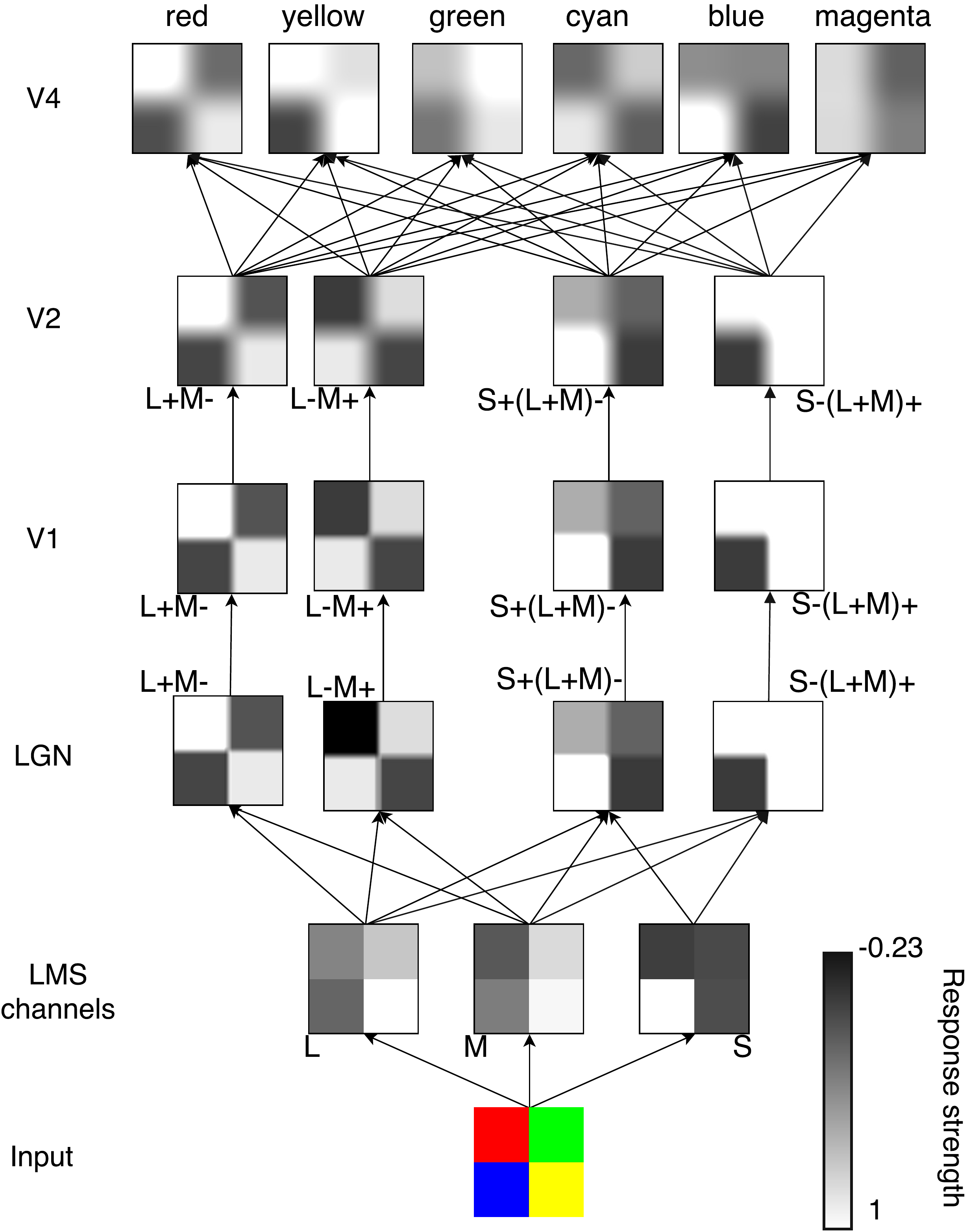}
	\caption{An example showing each layer of the hierarchical color model on an image with red, green, blue, yellow regions. The neuron type is written next to each square. First, the input image is converted into LMS channels. The channels are shown as L, M, and S, from left to right. Each square in the model layers represents an array of a neuron type. The receptive field of each neuron in these arrays is centered at the corresponding pixel location. The neuron responses are shown in grayscale, with minimum response as black, and maximum response as white. For example, in the array for neuron type $\text{L+M-}$, strong and moderate activities are observed for neurons with receptive fields inside the red and yellow regions. The dark lines around each neuron type activities are shown only for the purpose of this figure and are not parts of the activities.}
	\label{fig:rgby}
\end{figure}
In contrast, Figure~\ref{fig:rgby} depicts the network in action. That is, in this figure, each layer consists of a number of maps, each of which showing activities of neurons of a single type with their receptive fields centered at the corresponding pixel location in the image. For example, the map labeled as red in layer V4 shows the activities of model V4 neurons sensitive to the red hue with receptive fields centered at the corresponding pixels. 

Stacking the maps in the model V4 layer, in the order shown in Figure~\ref{fig:rgby}, will result in a three dimensional array. Each column of this array can be interpreted as a cluster of hue-selective patches, with neighboring patches sensitive to related hues, similar to those observed in V4 of monkeys reported in~\cite{li2014Map}. In other words, each column forms a rainbow of patches. Moreover, the neurons within each column, just like the patches observed in~\cite{li2014Map}, share the same local visual field and have largely overlapping local visual fields with those in their neighboring columns or clusters. In fact, these columns of color-sensitive neurons correspond to the hypothesized color columns of Barlow~\cite{Barlow1986}, which were later suggested to be the building units of glob cells in PIT~\cite{Conway_Tsao2009}.

In order to keep our model simple and avoid second-order equations, we skipped lateral connections between neuron types. However, these are part of future development of a second-order model.

\subsection{Model LGN Cells}
The first layer of the hierarchy models single-opponent LGN cells. The LGN cells are characterized by their opponent inputs from cones. For example, LGN cells receiving excitatory input from L cones and inhibitory signals from M cones are known as $\text{L+M-}$ cells. Model LGN cell responses were computed by~\cite{shapley2011color}:
\begin{eqnarray}
R_\text{LGN} & = \phi(& a_\text{L} (G(x, y, \sigma_\text{L}) \ast R_\text{L}) +\nonumber\\
			    &    & a_\text{M} (G(x, y, \sigma_\text{M}) \ast R_\text{M}) +\nonumber\\
			    &    & a_\text{S} (G(x, y, \sigma_\text{S}) \ast R_\text{S})), 
\label{eq:LGN}
\end{eqnarray}
where $\ast$ represents convolution. In this equation, model LGN response, $R_\text{LGN}$, is computed 
by first, linearly combining cone activities, $R_\text{L}$, $R_\text{M}$, and $R_\text{S}$, convolved with normalized Gaussian kernels, $G$, of different standard deviations, $\sigma$, followed by a linear rectification, $\phi$. For model LGN neurons, we set $\tau = -1$ and $s=1$ to ensure the responses of these neurons are linear combinations of the cone responses~\cite{DeValois2000physio,Lennie1990chromatic}. The differences in standard deviations of the Gaussian kernels ensure different spatial extents for each cone as described in~\cite{Reid2002}. Each weight in Eq.~\ref{eq:LGN}, determines presence/absence and excitatory/inhibitory effect of the corresponding cone. Following~\cite{Reid2002} and~\cite{johnson2004cone}, the weights used for model LGN cells are shown in Table~\ref{table:LGN_weights}. As an example, consider the weights for $\text{L+M-}$ cells. These neurons receive equal but opposite contributions from L and M cones, while S cones with weight 0 exhibit no contribution. Figure~\ref{fig:single_opp} depicts an example of the spatial profile of single-opponent $\text{L+M-}$ cells. In this example, L and M cones have excitatory and inhibitory effects respectively, while S cones are absent with no effects. In total, we modeled four 
different LGN neuron types, $\text{L+M-}$, $\text{L-M+}$, $\text{S+(L+M)-}$, and $\text{S-(L+M)+}$, which are known to best respond to reddish, cyan-like, lavender, and lime hues respectively~\cite{Conway2001spatial}. 
\begin{table*}[h]
\small\sf\centering
\captionsetup{justification=centering}
\caption{The choice of cone weights for model LGN cells.\label{table:LGN_weights}}
\begin{tabular}{cccc}
\hline
LGN neuron types&L cone weight&M cone weight&S cone weight\\
\hline
$\text{L+M-}$&1&-1&0\\
$\text{L-M+}$&-1&1&0\\
$\text{S+(L+M)-}$&-0.5&-0.5&1\\
$\text{S-(L+M)+}$&0.5&0.5&-1\\
\hline
\end{tabular}\\[10pt]
\end{table*}

\subsection{Model V1 and V2 cells}
Local hue in V1, as suggested in~\cite{shapley2002neural} and~\cite{johnson2004cone}, can be encoded by single-opponent cells. To obtain such a representation in model layers V1 and V2, the responses are determined by convolving input signals with a Gaussian kernel. Note that since single-opponency is implemented in model LGN layer, by simply convolving model LGN signals with a Gaussian kernel, we will also have single-opponency in model layers V1 and V2. The local hue responses of V1 and V2 were obtained by:
\begin{equation}
R_{\text{V1}} = \phi(G(x, y, \sigma_{\text{V1}}) \ast R_\text{LGN}),
\label{eq:V1_resp}
\end{equation}
and
\begin{equation}
R_{\text{V2}} = \phi(G(x, y, \sigma_{\text{V2}}) \ast R_{\text{V1}}),
\label{eq:V2_resp}
\end{equation}
where $\phi$ is the rectifier in Eq.~\ref{eq:rectifier}. With $\tau = 0$ and $s = 1$ for the rectifier, our model V1 and V2 neurons will be nonlinear functions of cone activations. In Eq.~\ref{eq:V1_resp}, substituting $R_\text{LGN}$ with any of the three model LGN neuron type responses will result in a corresponding V1 neuron type. This applies to model V2 neurons. That is, in Eq.~\ref{eq:V2_resp}, substituting $R_{\text{V1}}$ with each of the three model V1 neuron type responses will yield a similar model V2 neuron type. Therefore, there will be four neuron types in layers V1 and V2 corresponding to $\text{L+M-}$, $\text{L-M+}$, $\text{S+(L+M)-}$, and $\text{S-(L+M)+}$.

The size of the Gaussian kernels for each of these neurons simply determines their receptive field sizes. In our implementation, the receptive field size doubles from one layer to the next following a similar observations in the ventral stream~\cite{freeman2011metamers}.

\subsection{Model V4 cells.}
We modeled V4 neurons representing local hue using a weighted sum of convolutions from the four model V2 neuron types. More specifically, V4 responses are computed as:
\begin{eqnarray}
R_{\text{V4}} &= \phi(& a_r (G(x, y, \sigma_{\text{V4}}) \ast R_{\text{V2, r}}) +\nonumber\\
                       &  & a_g (G(x, y, \sigma_{\text{V4}}) \ast R_{\text{V2, g}}) +\nonumber\\
                       &  & a_y (G(x, y, \sigma_{\text{V4}}) \ast R_{\text{V2, y}}) +\nonumber\\
                       &  & a_b (G(x, y, \sigma_{\text{V4}}) \ast R_{\text{V2, b}})),
\label{eq:V4}
\end{eqnarray}
where $R_{\text{V2, r}}$, $R_{\text{V2, g}}$, $R_{\text{V2, y}}$, and $R_{\text{V2, b}}$ are responses of V2 channels corresponding to $\text{L+M-}$, $\text{L-M+}$, $\text{S-(L+M)+}$, and $\text{S+(L+M)-}$ respectively. For notation simplicity, we are referring to these neuron types as r, g, y, and b, even though they do not correspond to red, green, yellow, and blue colors as discussed earlier. Again, $\phi$ is the rectifier introduced in Eq.~\ref{eq:rectifier}, with $\tau = 0$ and $s = 1$.

In Equation~\ref{eq:V4}, the weights $a_r$, $a_g$, $a_y$, $a_b$ determine the hue to which a model V4 neuron shows selectivity. For example, for the setting $a_r = 1, a_g=0, a_y=0, a_b=0$, the V4 neurons show highest activity to reddish hues, as $\text{V2, r}$ encodes this hue, while the setting $a_r = 0, a_g=1, a_y=0, a_b=0$ results in neurons selective to cyan-like hues. 

\begin{figure}[h!]
	\centering
	\includegraphics[width=0.48\textwidth]{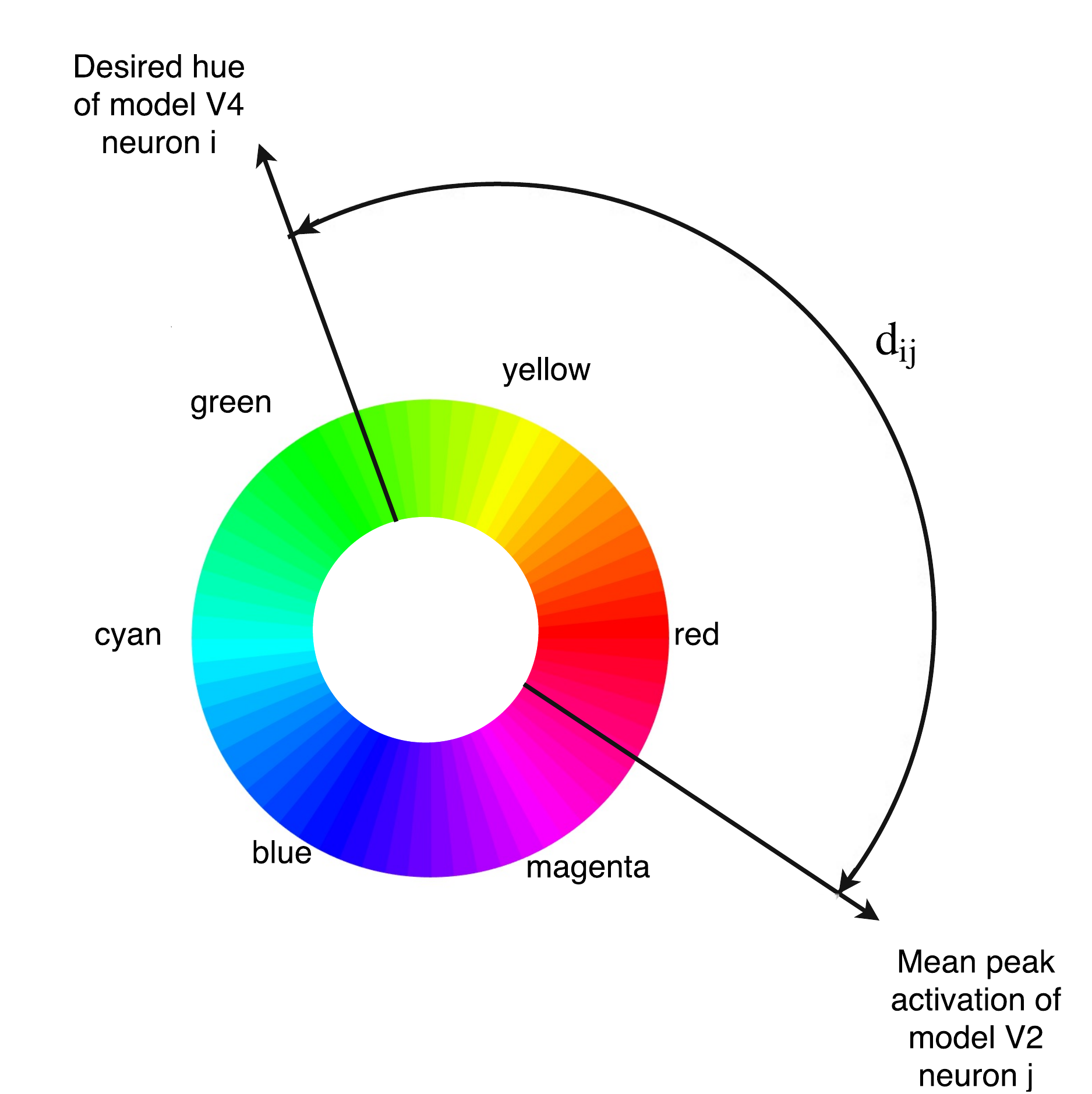}
	\caption{Hues on a circle. The red hue at 0 deg starts from 3 O'clock on the circle. The main hues are 60 deg apart on this circle. The two straight arrows show examples of hues for which the mean peak activations of an imaginary model V2 neuron occurs and the desired hue of a model V4 neuron. The arc shows the hue angle, or the angular distance, between the two hues on the circle, represented as $d_{ij}$. This hue distance determines the weight of model V2 neuron $j$ to model V4 neuron $i$. }
	\label{fig:hue_wheel}
\end{figure}

In model layer V4, we implemented six different neuron types according to distinct hues: red, yellow, green, cyan, blue, and magenta. The chosen hues are 60 deg apart on the hue circle shown in Figure~\ref{fig:hue_wheel}, with red at 0 deg. These hues were also identified on the V4 color map study~\cite{li2014Map}. From here on, we will refer to these neurons based upon their selectivities, e.g., model V4 red or model V4 cyan neurons. Although here we limit the number of modeled neuron types in this layer to six, we would like to emphasize that changes in combination weights will lead to neurons with various hue selectivities in this layer. Modeling neurons with selectivities to a wide variety of hues with yet narrower tunings could be accomplished in higher layers, such as IT, by combining hue-selective model neurons in V4. 

In order to determine the weights from V2 to V4 neurons, we considered the distance between mean peak activations of model V2 neurons to the desired hue in a model V4 cell. For example, consider the two arrows on the hue circle shown in Figure~\ref{fig:hue_wheel}. One specifies the hue at which the mean peak activations of the model V2 neuron $j$ occurs, while the other is at the desired hue for the model V4 neuron $j$. The hue angle between these two hues on the circle is represented by $d_{ij}$. Then, the weight $w_{ij}$ from model V2 neuron $j$ to model V4 neuron $i$ is determined by:
\begin{equation}
w_{ij} = \frac{\mathcal{N}(d_{ij}; 0, \sigma)}{Z_i},
\label{eq:V4_weights}
\end{equation}
where $\mathcal{N}(.;0, \sigma)$ represents a normal distribution with 0 mean and $\sigma$ standard deviation, and $Z_i$ is a normalizing constant obtained by
\begin{equation}
Z_i = \sum_{j=1}^4\mathcal{N}(d_{ij}; 0, \sigma).
\end{equation}
The weights used for each of V4 neuron types are summarized in Table~\ref{table:V4_weights}. For example, the red V4 neuron activity comes mainly from the $L+M-$ neurons in V2, and the weight from $L-M+$ is negligible compared to the rest of the weights. This is not surprising as previous research by Webster~\etal~\cite{Webster2000variations} found unique red in human subjects has largest contributions from L cones. As another example, the magenta V4 neuron responses are mainly contributed by activities of $S+(L+M)-$ and $L+M-$ neurons in V4, \ie, lavender and reddish signals.
\begin{table*}[t]
\small\sf
\centering
\captionsetup{justification=centering}
\caption{The choice of relative weights used for model V4 cells.\label{table:V4_weights}}
\begin{tabular}{ccccc}
\hline
V4 neuron types&$\text{L+M-}$ weight&$\text{L-M+}$ weight&$\text{S+(L+M)-}$ weight&$\text{S-(L+M)+}$ weight\\
\hline
red&0.85636&0.00028984&0.041238&0.10211\\
\hline
yellow&0.38019&0.022312&0.0005716&0.59692\\
\hline
green&0.031002&0.31546&0.012604&0.64093\\
\hline
cyan&0.00038727&0.68329&0.2109&0.10543\\
\hline
blue&0.014012&0.29034&0.69225&0.0034021\\
\hline
magenta&0.37948&0.031779&0.58531&0.0034362\\
\hline
\end{tabular}\\[10pt]
\end{table*}

In Figure~\ref{fig:rgby}, at the V4 layer, from left to right, the neurons selective to red, yellow, green, cyan, blue, and magenta are displayed. As expected, model V4 magenta neurons, for instance, show activations across both red and blue regions of the stimulus.

\section{Results}
In this section, we explain three sets of experiments performed with our model. First, we study the activations of model V4 neurons to various hues and make a comparison with biological V4 neurons. Second, we show that given the model V4 activations, any hue in the HSL space can be reconstructed. Similar to the stimuli set designed and employed by~\cite{li2014Map}, our stimuli were hues from the HSL space. This choice of the color space for experiments was made to enable easy comparison.

\subsection{Model Neuron Selectivities}
\begin{figure}[h!]
	\centering
	\includegraphics[width=0.47\textwidth]{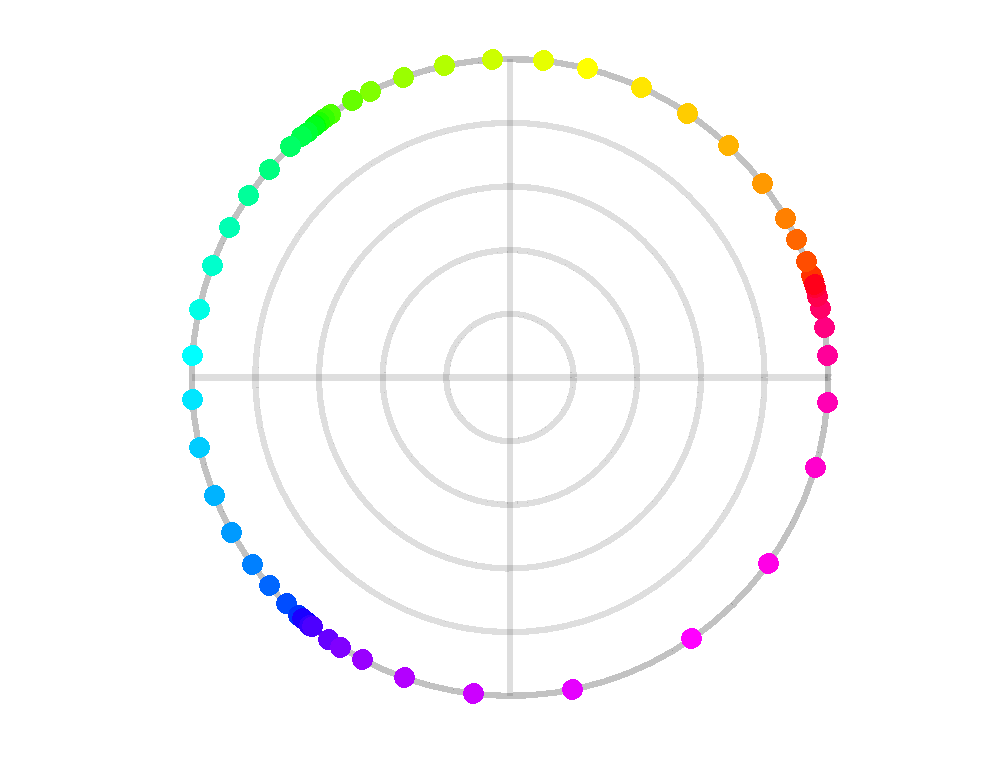}
	\caption{The 60 sampled hues from the HSL space mapped onto a unit circle in the MacLeod and Boyton space. As shown, the hues are not uniformly spaced on the unit circle. The hues are rotated on the unit circle, with different hue angles, compared to their corresponding angles on the hue circle in Figure~\ref{fig:hue_wheel}. The four cardinal axes in this space correspond to reddish, lime, cyan-like and lavender hues.}
	\label{fig:hues_in_MBspace}
\end{figure}

In this experiment, in order to test the effectiveness of our approach for modeling local hues, we examined the peak selectivity of each hue-selective neuron in individual layers of our network. For this purpose, we sampled the hue dimension of the HSL space. We keep saturation and lightness values constant and set to 1 and 0.5 respectively. Our sampling consists of 60 different hues in the range of $[0, 360)$ degrees, separated by 6 degrees. 
Figure~\ref{fig:hues_in_MBspace} presents the 60 sampled hues on a unit circle in the MacLeod and Boyton diagram. All plots, in what follows, demonstrate the tuning curves in the MacLeod and Boyton space, even though the stimuli are sampled from HSL. 
As follows from Figure~\ref{fig:hues_in_MBspace}, these hues are not uniformly spaced on the unit circle and are rotated with respect to their corresponding hue angle on the hue circle in Figure~\ref{fig:hue_wheel}. For example, the hue at 0 degree on the hue circle in Figure~\ref{fig:hue_wheel} is at about 18 degree on the unit circle in the MacLeod and Boyton diagram.

It is worth mentioning here that in determining the weights from V2 to V4 neurons, one might take the distances of V2 peak responses and desired hue selectivity of V4 neurons on the unit circle in the MacLeod and Boyton space. Alternatively, these distances could be measured based on the corresponding hue values in the HSL space representation. We experimented with both settings and did not observe any major differences in results. What we report here will be based on the V4 weights computed in the latter setting.

\begin{figure*}[h!]
	\centering
	\subfigure{\includegraphics[width=0.35\textwidth]{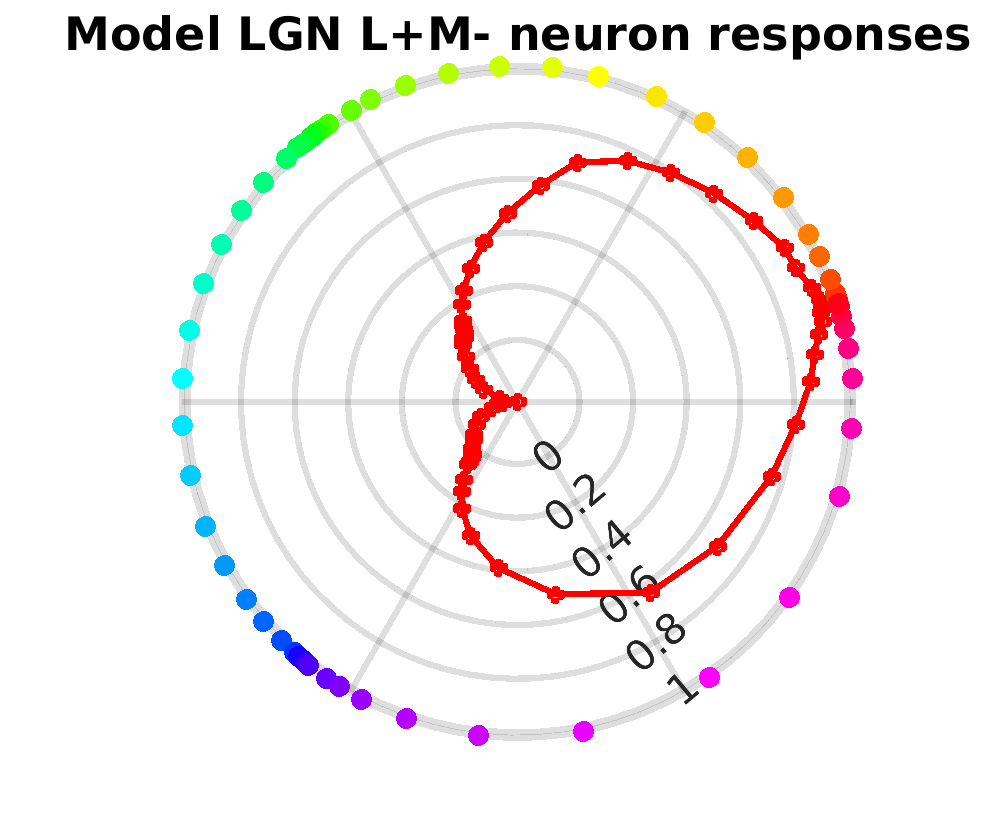}}~
	\subfigure{\includegraphics[width=0.35\textwidth]{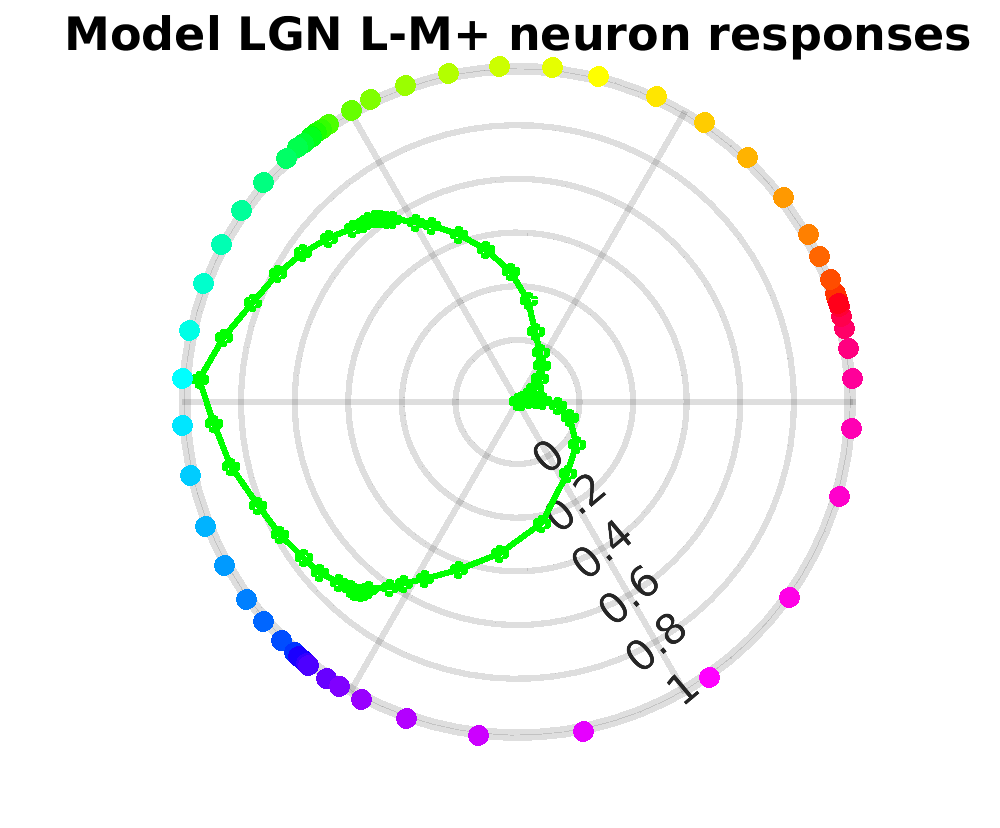}}\\
	\subfigure{\includegraphics[width=0.35\textwidth]{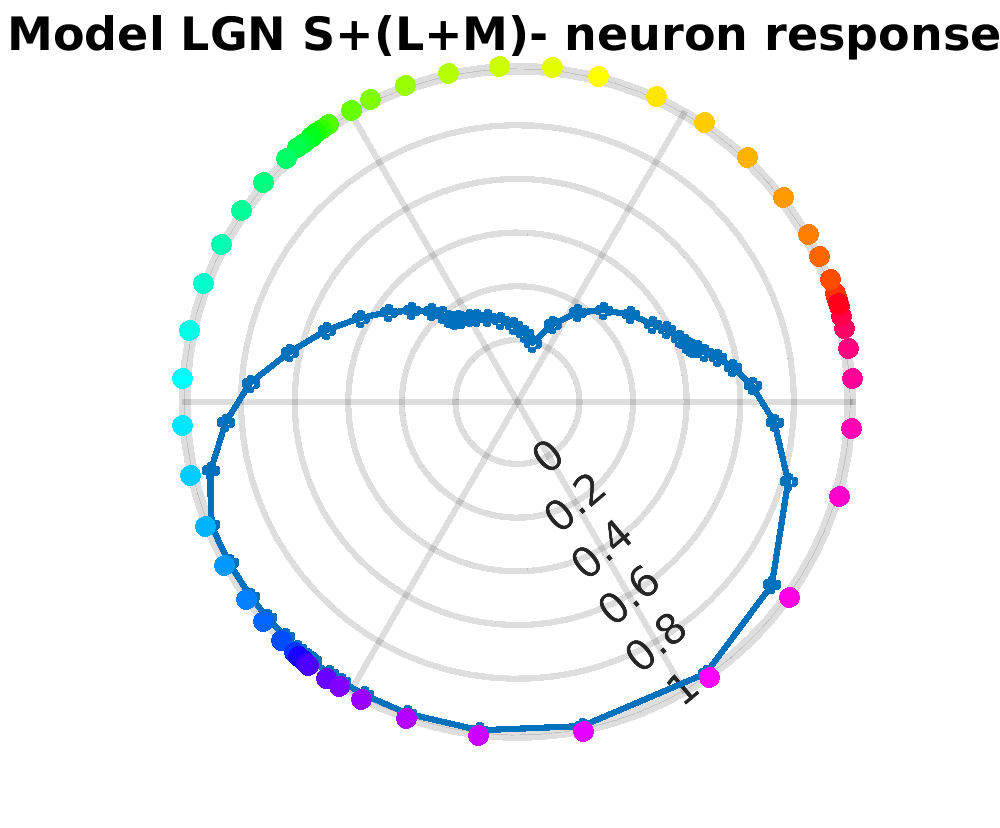}}~
	\subfigure{\includegraphics[width=0.35\textwidth]{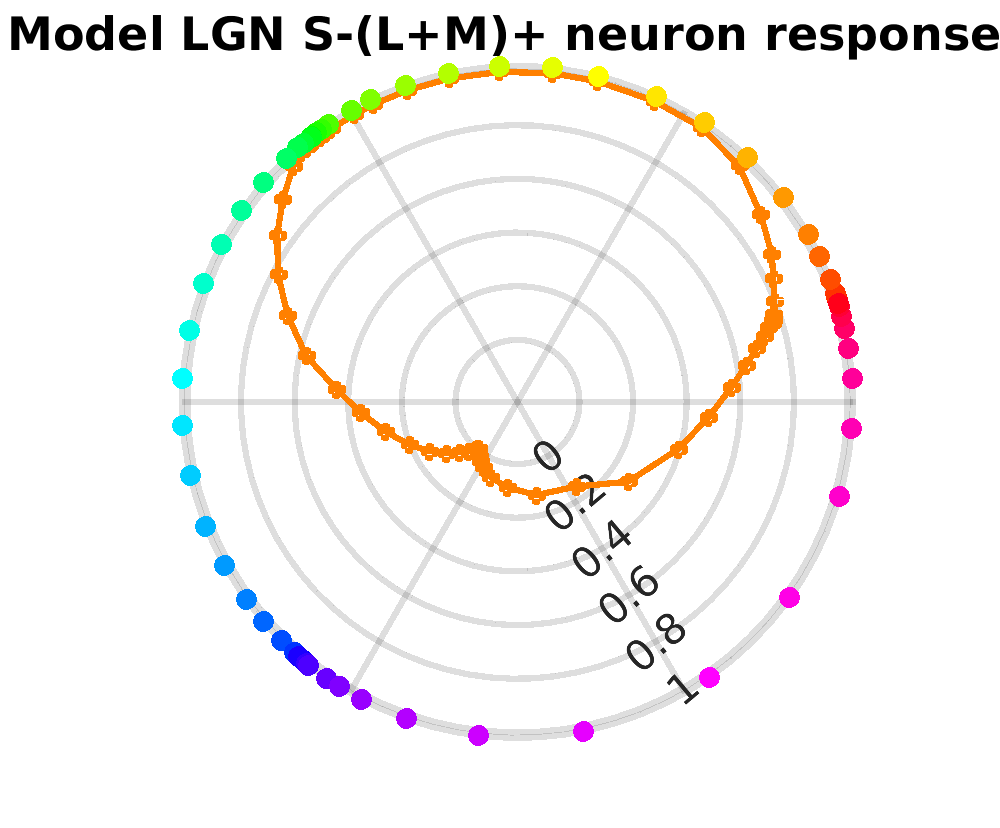}}\\
	\caption{Model LGN neuron responses to hues sampled from the hue dimension in HSL space. The sampled hues are 6 degrees apart. In each plot, the angular dimension shows the hues in the MacLeod and Boyton diagram, and the radial dimension represents the response level of the neuron. Model V2 neurons from top to bottom and left to right: $L+M-$, $L-M+$, $S+(L+M)-$, $S-(L+M)+$.}
	\label{fig:hue_resp_LGN}
\end{figure*}

\begin{figure*}[h!]
	\centering
	\subfigure{\includegraphics[width=0.37\textwidth]{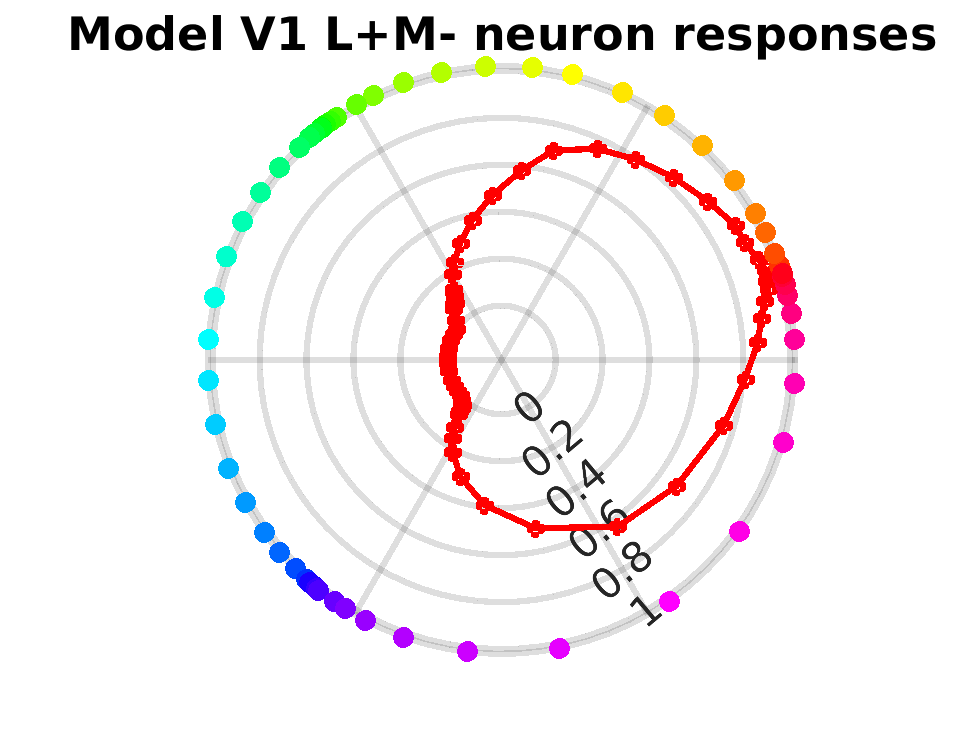}}~
	\subfigure{\includegraphics[width=0.35\textwidth]{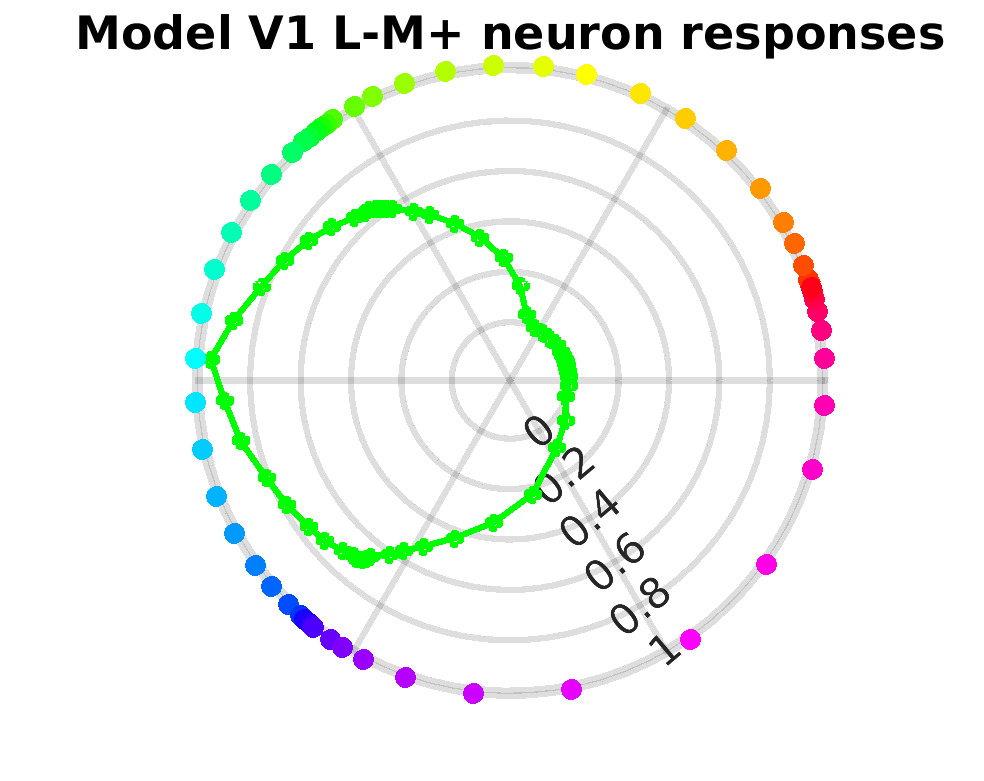}}\\
	\subfigure{\includegraphics[width=0.39\textwidth]{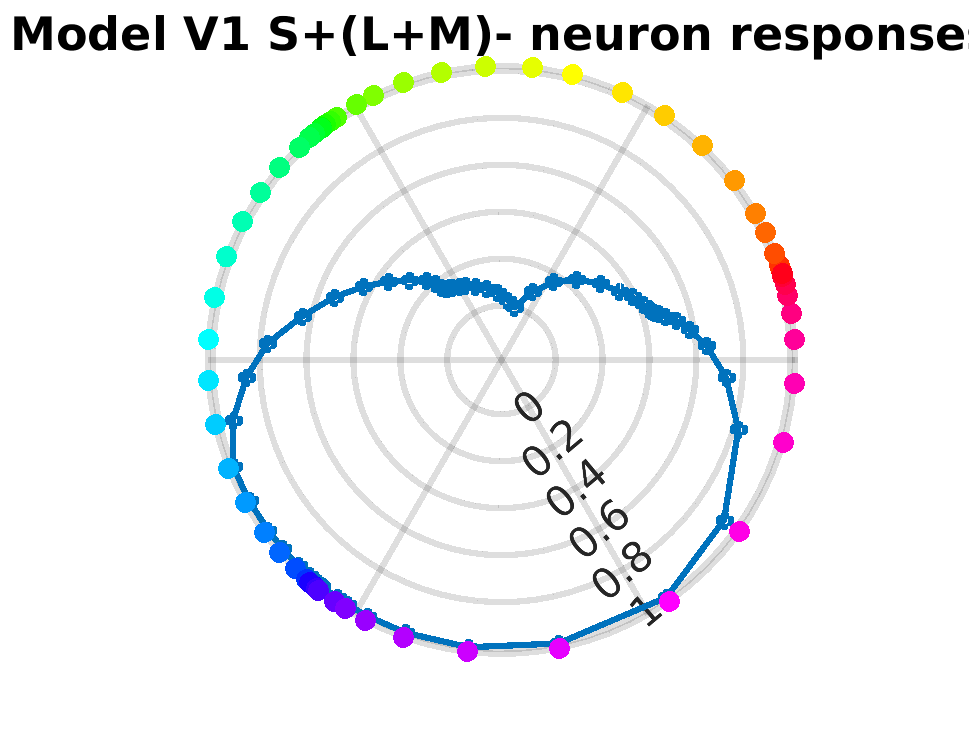}}~
	\subfigure{\includegraphics[width=0.35\textwidth]{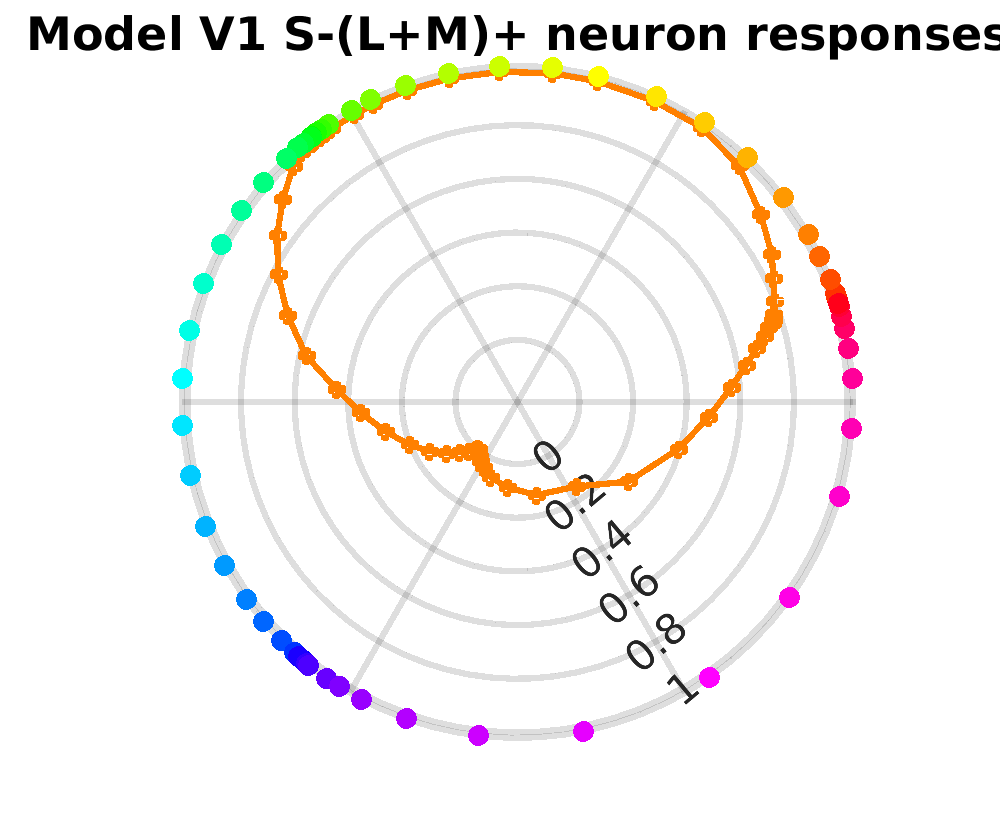}}\\
	\caption{Model V1 neuron responses to hues sampled from the hue dimension in HSL space. The sampled hues are 6 degrees apart. In each plot, the angular dimension shows the hue angles in the MacLeod and Boyton diagram, and the radial dimension represents the response level of the neuron. Model V1 neurons from top to bottom and left to right: $L+M-$, $L-M+$, $S+(L+M)-$, $S-(L+M)+$.}
	\label{fig:hue_resp_V1}
\end{figure*}

\begin{figure*}[h!]
	\centering
	\subfigure{\includegraphics[width=0.35\textwidth]{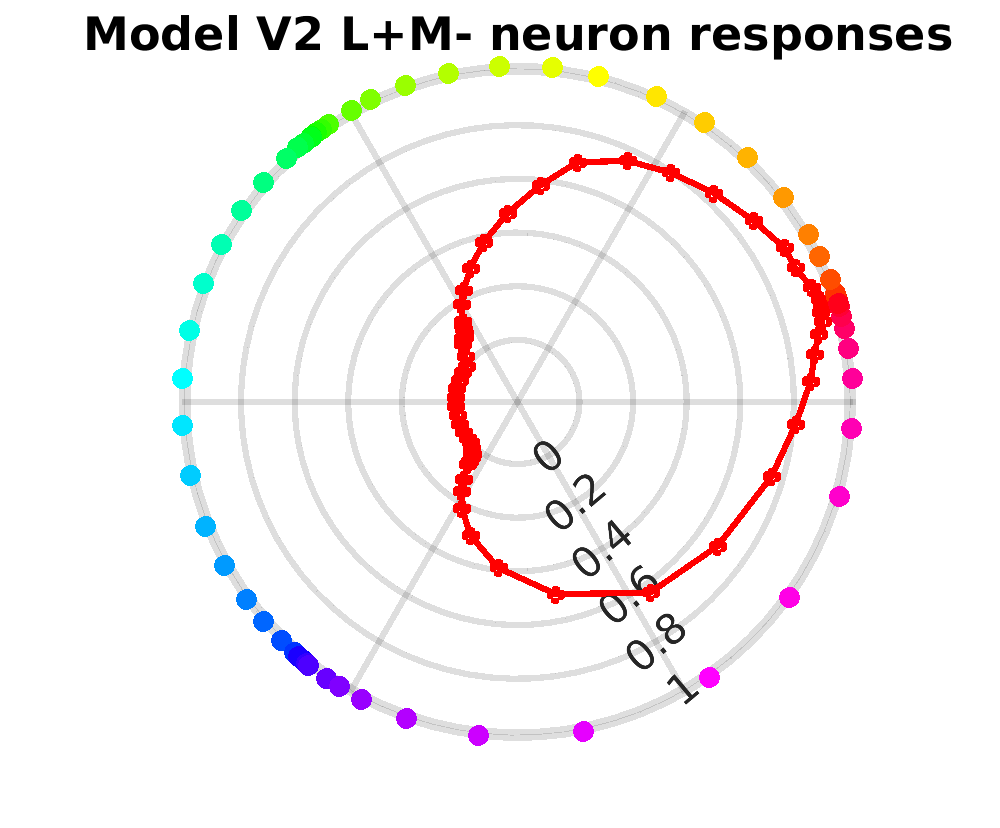}}~
	\subfigure{\includegraphics[width=0.35\textwidth]{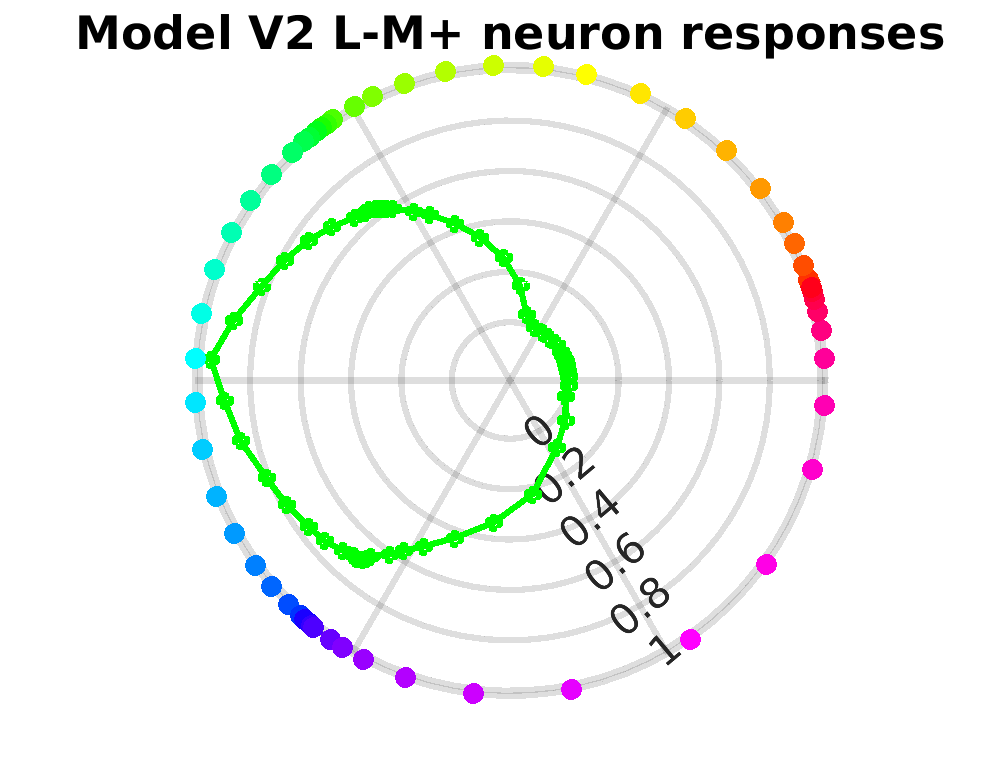}}\\
	\subfigure{\includegraphics[width=0.35\textwidth]{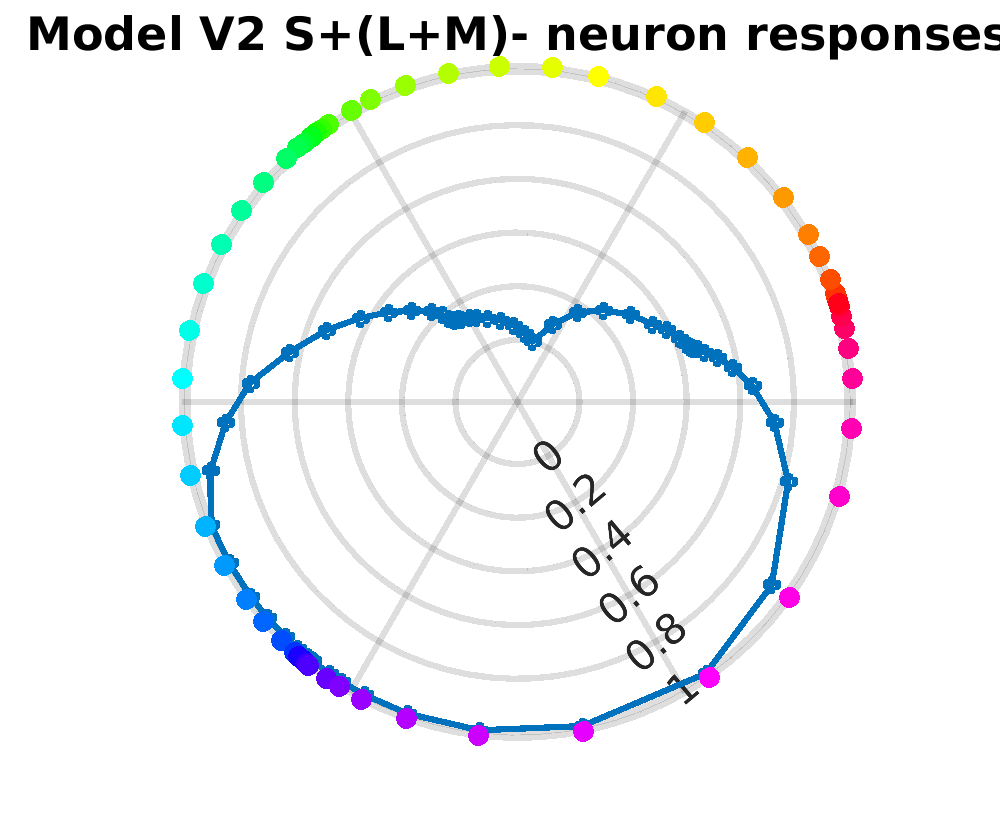}}~
	\subfigure{\includegraphics[width=0.35\textwidth]{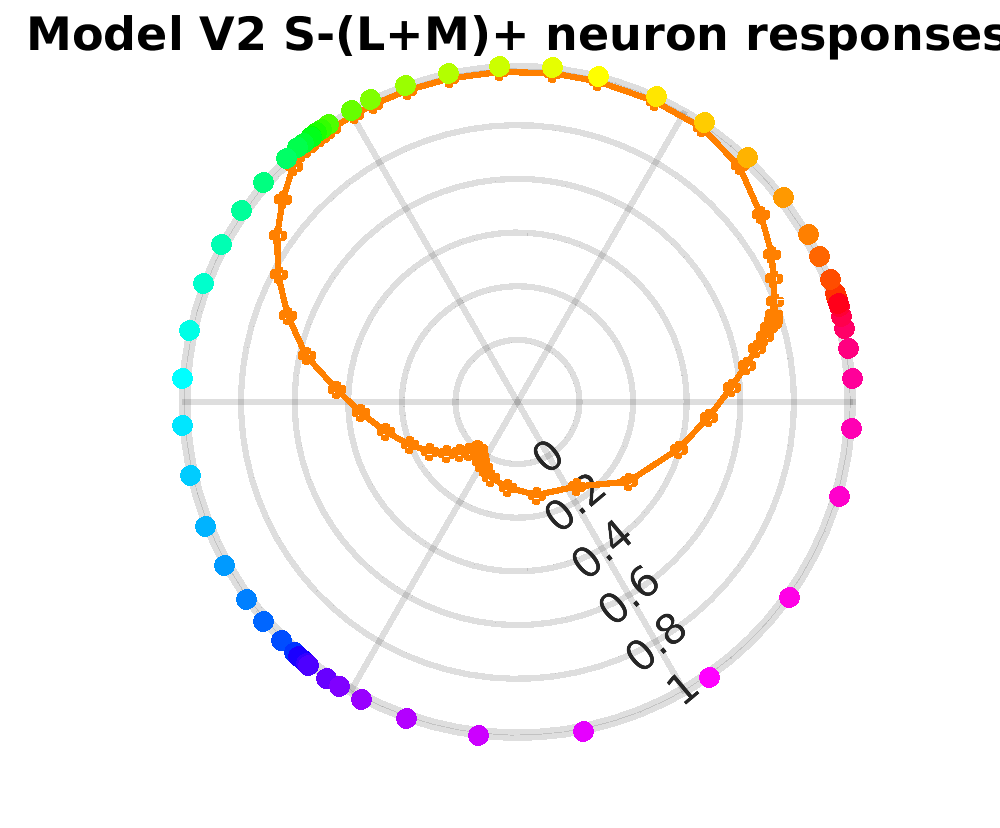}}\\
	\caption{Model V2 neuron responses to hues sampled from the hue dimension in HSL space. The sampled hues are 6 degrees apart. In each plot, the angular dimension shows the hue angles in the MacLeod and Boyton diagram, and the radial dimension represents the response level of the neuron. Model V2 neurons from top to bottom and left to right: $L+M-$, $L-M+$, $S+(L+M)-$, $S-(L+M)+$. The tuning curves of V2 neurons look identical to those of model V1 cells (see text for details).}
	\label{fig:hue_resp_V2}
\end{figure*}

\begin{figure*}[h!]
	\centering
	\subfigure{\includegraphics[width=0.35\textwidth]{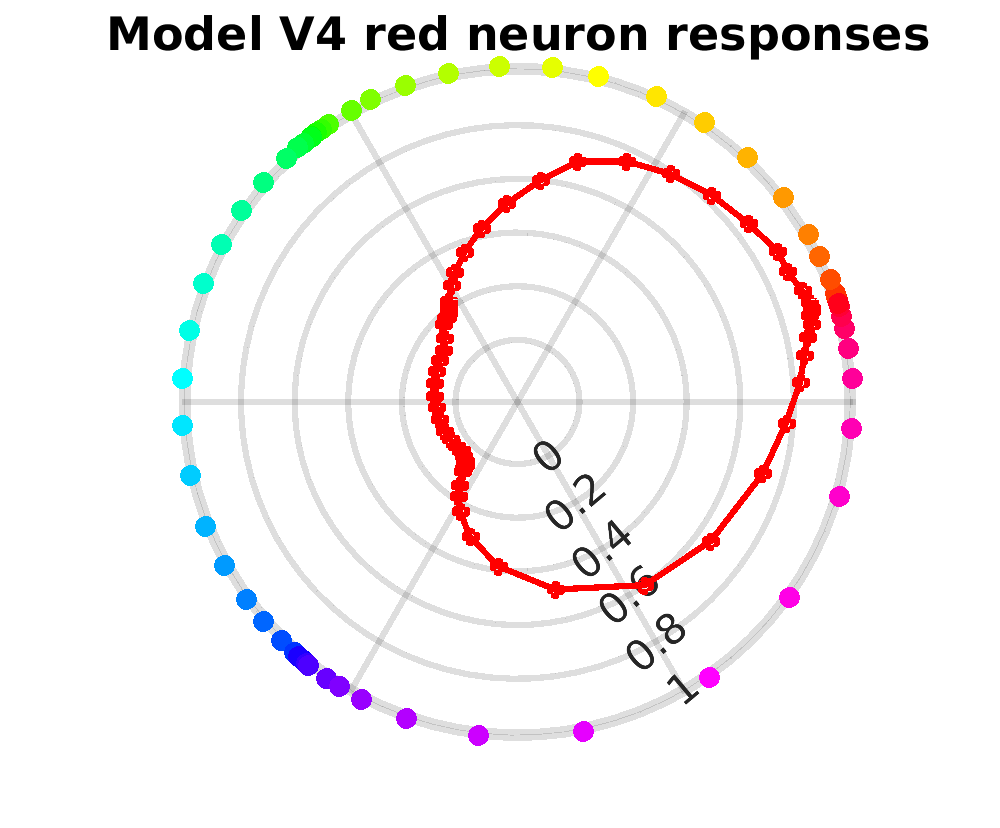}}~
	\subfigure{\includegraphics[width=0.35\textwidth]{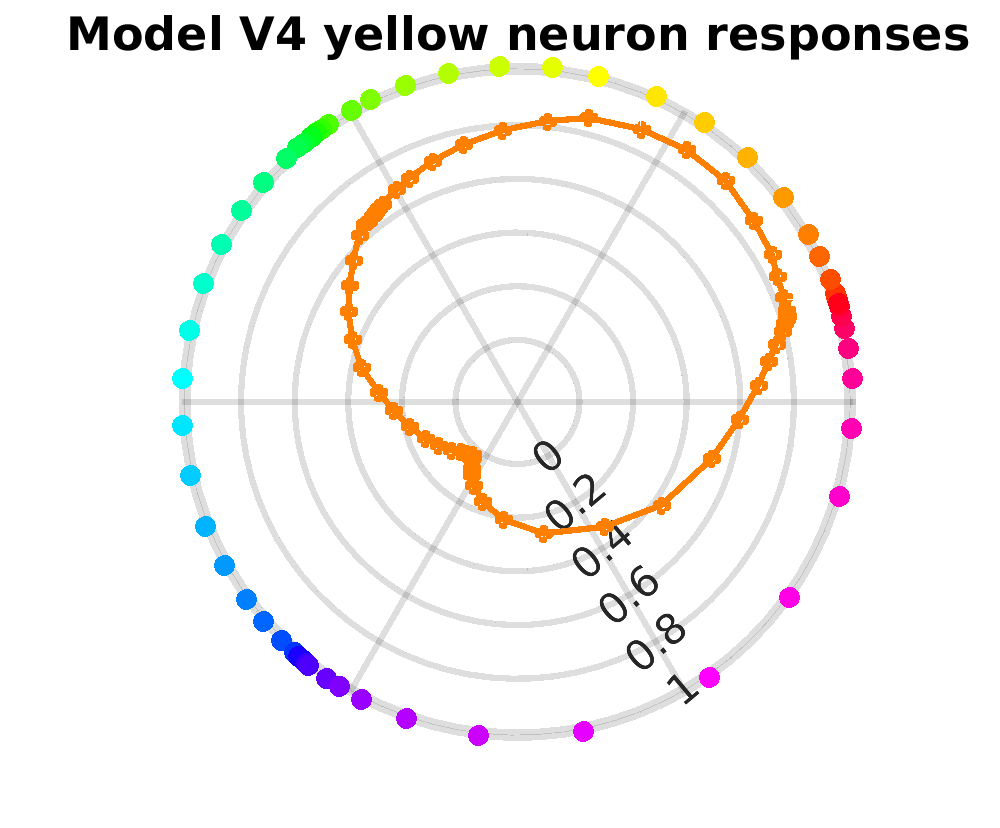}}\\
	\subfigure{\includegraphics[width=0.35\textwidth]{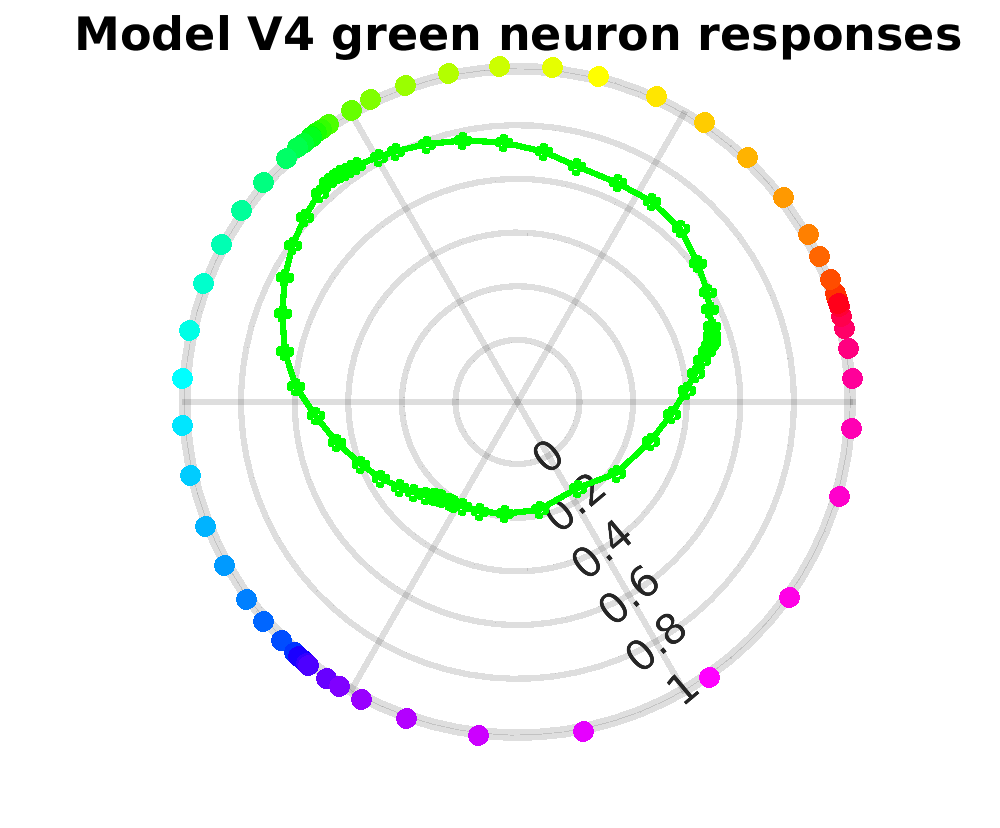}}~
	\subfigure{\includegraphics[width=0.36\textwidth]{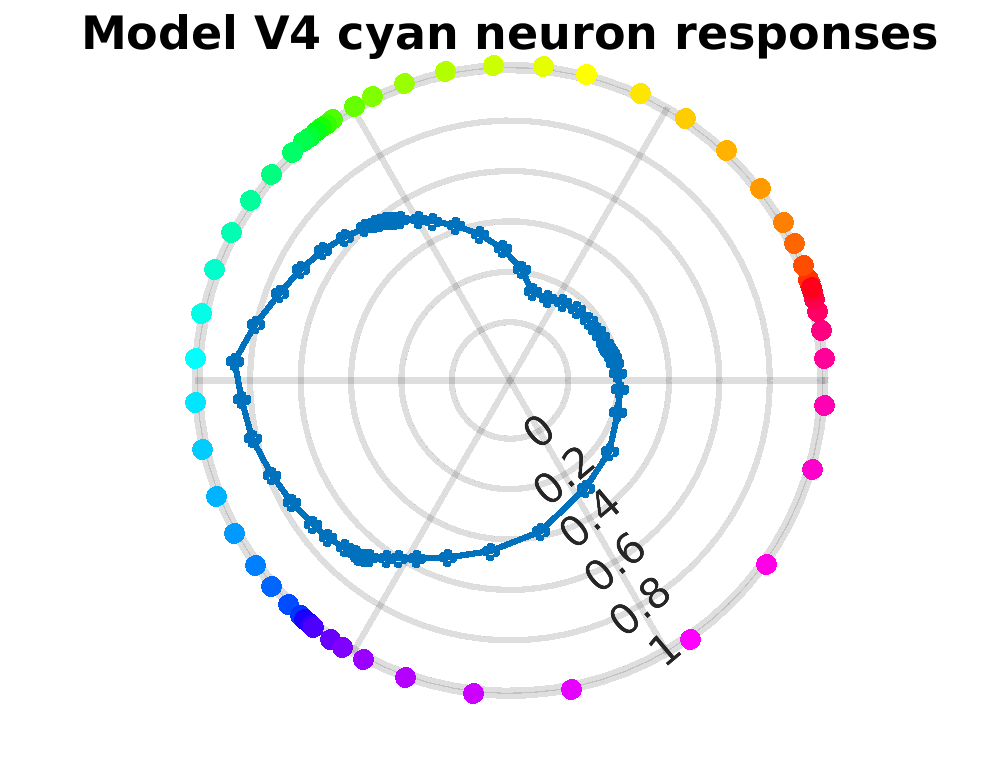}}\\
	\subfigure{\includegraphics[width=0.35\textwidth]{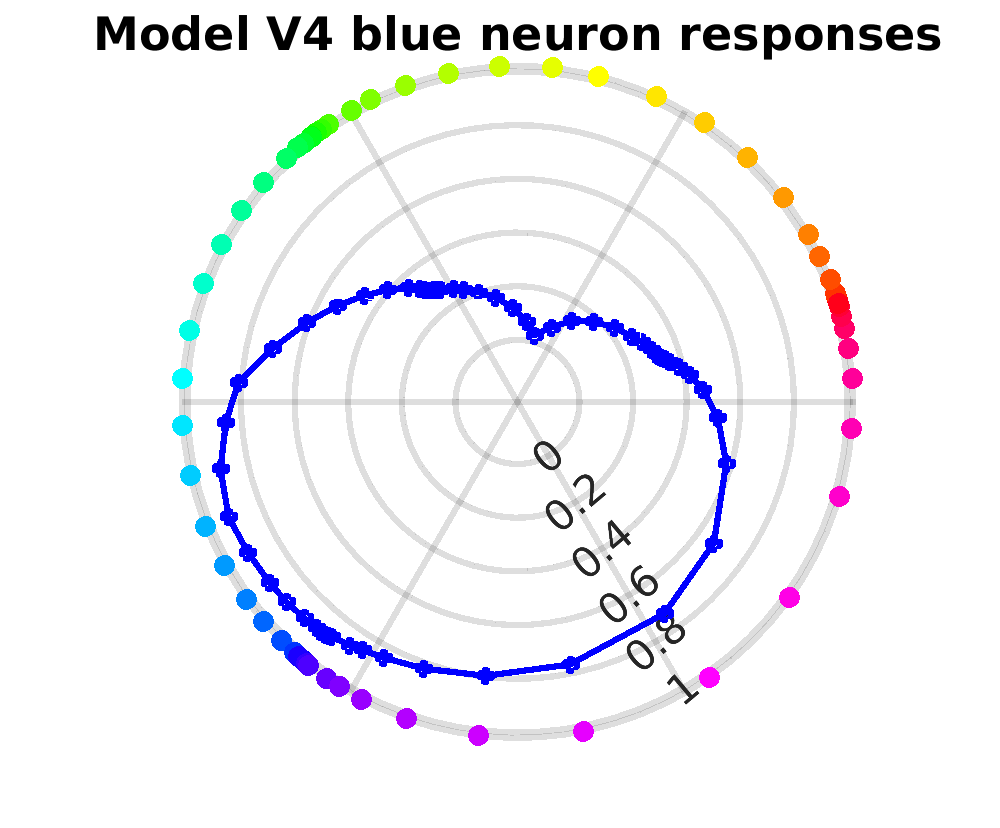}}~
	\subfigure{\includegraphics[width=0.34\textwidth]{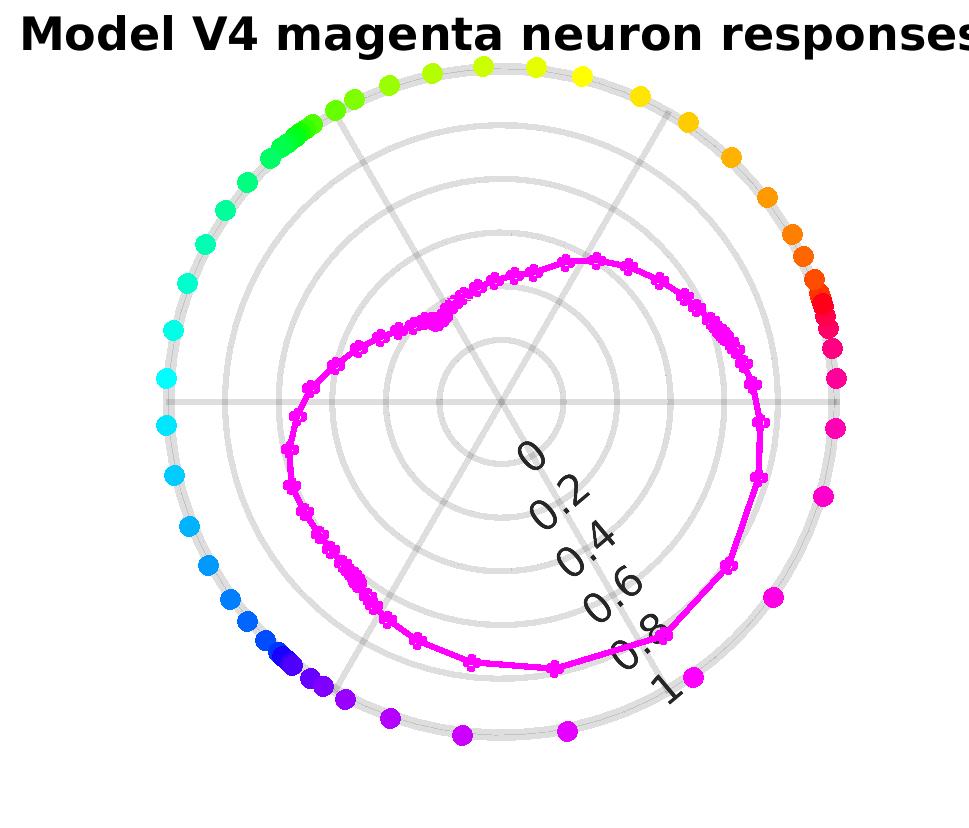}}\\
	\caption{Model V4 neuron responses to hues sampled from the hue dimension in HSL space. The sampled hues are 6 degrees apart. In each plot, the angular dimension shows the hue angles in the MacLeod and Boyton diagram}, and the radial dimension represents the response level of the neuron. Model V4 neurons from top to bottom and left to right: red, yellow, green, cyan, blue, and magenta.
	\label{fig:hue_resp_plots}
\end{figure*}

We present each of these 60 hues to the model and record the activities of model LGN, V1, V2 and V4 neurons. Plots in Figures~\ref{fig:hue_resp_LGN}, \ref{fig:hue_resp_V1}, \ref{fig:hue_resp_V2},~\ref{fig:hue_resp_plots} show model LGN, V1, V2, and V4 neuron activities to each of the sampled hues. 
In each plot, the circular dimension represents the hue angle in the MacLeod and Boyton diagram, and the radial dimension represents the response level of the neuron. Figures~\ref{fig:hue_resp_LGN},~\ref{fig:hue_resp_V1}, and \ref{fig:hue_resp_V2} depict the selectivity of single-opponent neurons. While the responses of $S+(L+M)-$ and $S-(L+M)+$ neurons in LGN, V1, and V2 layers look relatively similar, the differences due to the nonlinearity of V1 and V2 neurons imposed by the rectifier are obvious in the responses of $L+M-$ and $L-M+$. These plots also show higher sensitivities in $L+M-$ and $L-M+$ compared to those of $S+(L+M)-$ and $S-(L+M)+$. A closer look at Figures~\ref{fig:hue_resp_V1} and \ref{fig:hue_resp_V2} reveals that the tuning curves of model V1 and V2 neurons look identical. This effect is due to applying the rectifier to model V1 responses. Specifically, the responses of model V1 neurons are positive after rectification, which results in V2 neurons responses from Equation~\ref{eq:V2_resp} being simply a Gaussian-weighted combination of those of V1 cells. In fact, the average difference of responses between model V1 and V2 cells to the 60 tested hues were $0.3450\times10^{-6}$, $0.3533\times10^{-6}$, $0.5033\times10^{-6}$, $0.5535\times10^{-6}$ for the four single-opponent $L+M-$, $L-M+$, $S+(L+M)-$, $S-(L+M)+$ cells.

\begin{figure}[h!]
	\centering
	\includegraphics[width=0.48\textwidth]{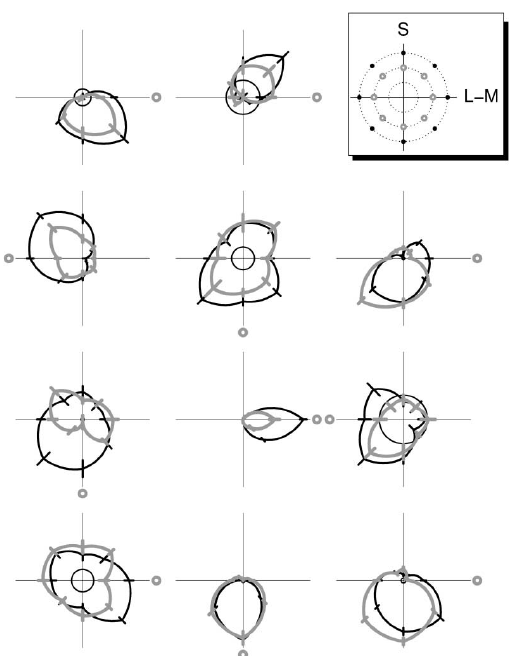}
	\caption{The tuning curves of V1 neurons studied for the effect of background chromacity by Wachtler~~\protect\etal~\protect\cite{Wachtler2003}. Each diagram shows the tuning curves for stimuli on gray background as a solid black curve. These neurons display an obvious nonlinearity in terms of cone inputs. Moreover, most of the neurons have wide tuning curves (figure adapted from~\protect\cite{Wachtler2003}).}
	\label{fig:Wachtler_V1_tunings}
\end{figure}

The tuning curves of our model V1 neurons show similarities to those of Macaque reported by Wachtler~\etal~\cite{Wachtler2003} as depicted in Figure~\ref{fig:Wachtler_V1_tunings}. The plots in both Figures~\ref{fig:hue_resp_V1} and \ref{fig:Wachtler_V1_tunings} show wide tunings with nonlinearity in terms of cone inputs.
It is also worth noting that $S+(L+M)-$ and $S-(L+M)+$ neurons in these early layers, with such wide tuning curves, do not encode unique hues. However, the tuning curves for $L+M-$ and $L-M+$ are narrower and peak around angles for reported unique hues~\cite{Webster2000variations}. Perhaps this could explain the disagreements in previous studies in which some suggested neurons in V1 are tuned to perceptual hues~\cite{DeValois2000physio}, and some rejected this suggestion~\cite{Webster2000variations}\cite{Wuerger2005} claiming higher order mechanisms are required for encoding of unique hues.

\begin{figure*}[!t]
	\centering
	\includegraphics[width=0.4\textwidth]{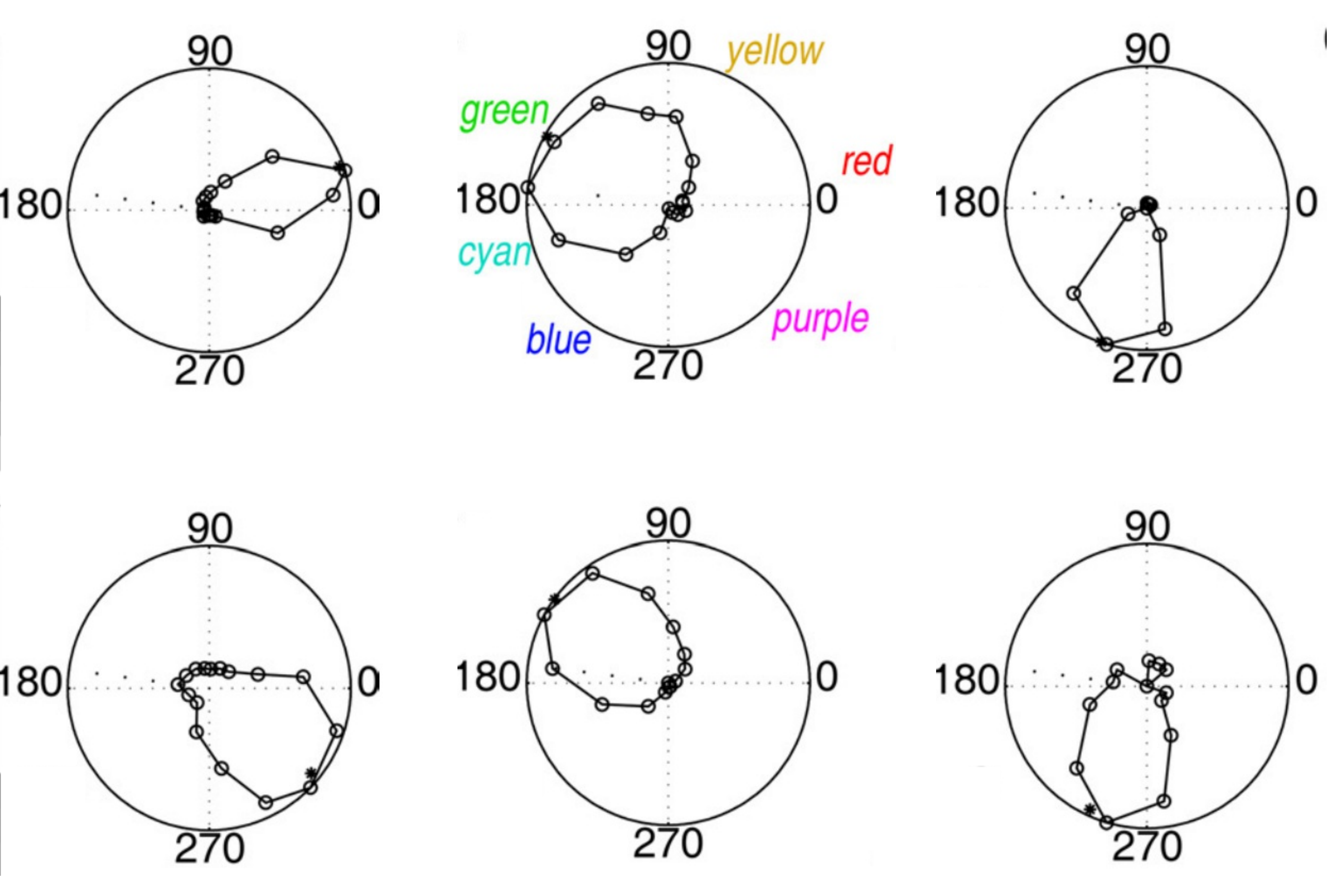}
	\caption{Tuning curves for 6 globs cells showing preference to red (left column), green (middle column) and blue (right column) colors, reported by Conway and Tsao~\protect\cite{Conway_Tsao2009}. The neurons have narrower tuning curves compared to those of V1 (figure adapted from~\protect\cite{Conway_Tsao2009}).}
	\label{fig:Conway_glob_tunings}
\end{figure*}

The plots in Figure~\ref{fig:hue_resp_plots}, for all model V4 neurons, show a peak of activity at their selectivity. The activities then start to diminish as the hues become more different from their selectivities. At the most distant hue, the hue 180 degrees apart from their selectivity, these neurons show almost no or close to zero activities. Comparing the tuning curves of V4 neurons with those of LGN and V2 shows that these neurons become more sensitive to their desired hues, with narrower tuning curves, as found by~\cite{Bohon2016,Zeki1980}. A qualitative comparison with the tuning curves of six glob cells in V4 selective to red, green and blue hues shown in Figure~\ref{fig:Conway_glob_tunings}, studied by Conway and Tsao~\cite{Conway_Tsao2009}, demonstrates a similar effect in both model and biological cells. This observation confirms that, by combining V4 neurons, even more sensitive neurons with more various hue selectivities can be modeled in higher layers such as IT.

\begin{figure*}[h!]
	\centering
	\subfigure[An example with various hues as bars]{\includegraphics[width=0.45\textwidth]{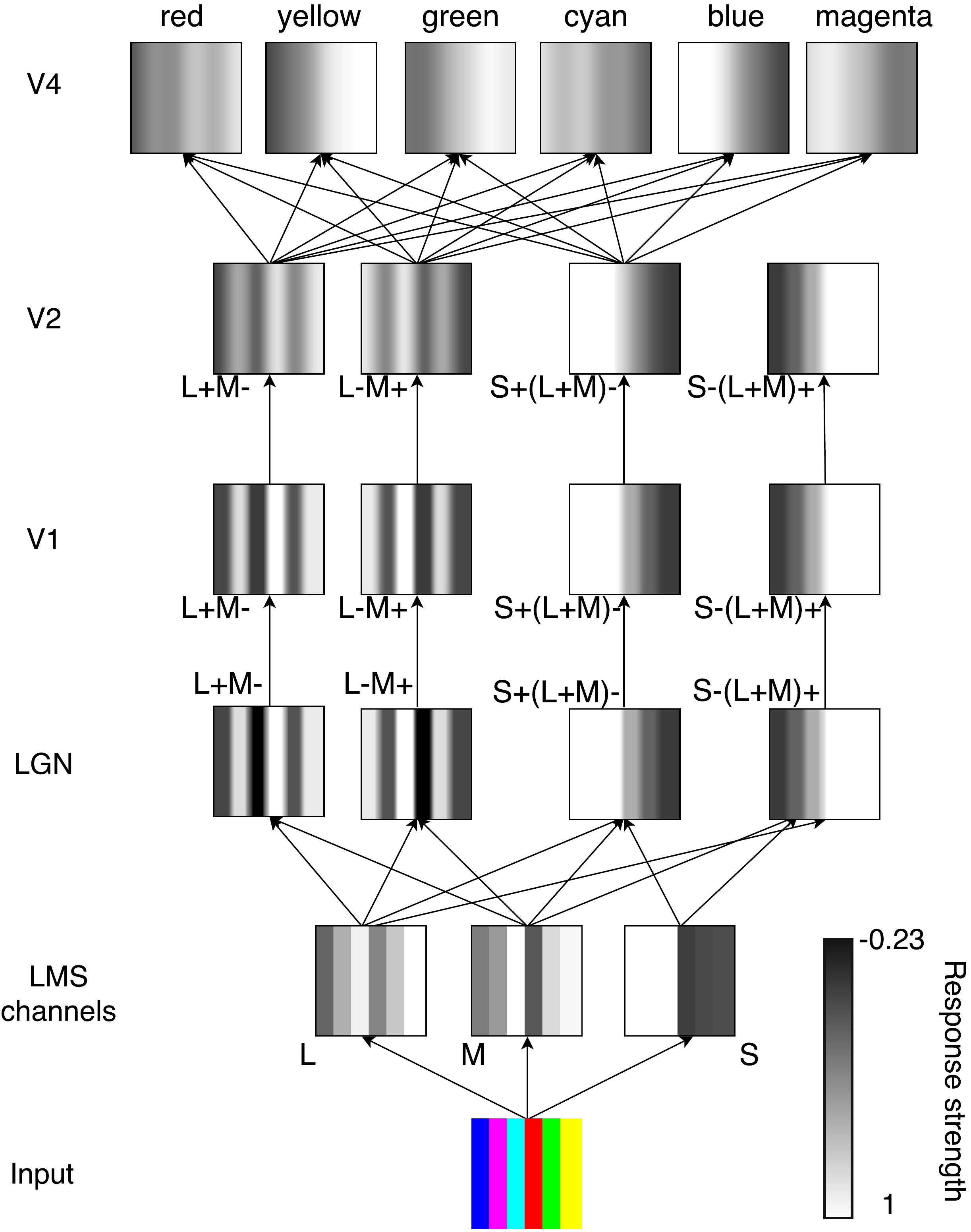}}~
	\subfigure[Example of hue circle as stimulus]{\includegraphics[width=0.45\textwidth]{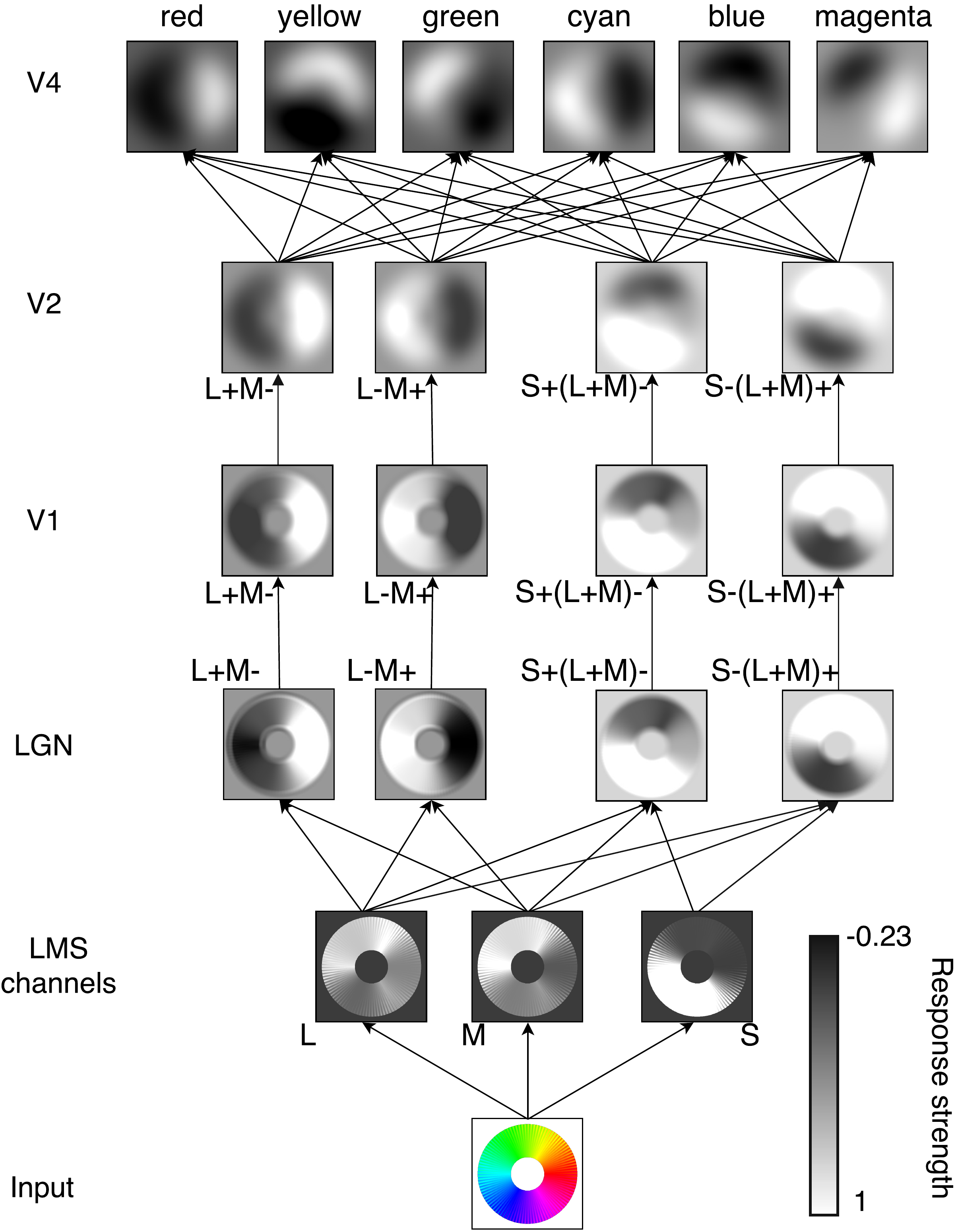}\label{subfig:ccm_result}}\\
	\caption{Qualitative examples of the hierarchical model. First, the input image is converted into LMS channels. The channels are shown as L, M, and S, from left to right. Each square in the model layers represents a map of neurons of a single type, with the neuron type written next to it. The receptive field of each neuron in these arrays is centered at the corresponding pixel location. The neuron responses are shown in grayscale, with minimum response as black, and maximum response as white. The dark lines around each neuron type activities are shown only for the purpose of this figure and are not parts of the activities.}
	\label{fig:qual_examples}
\end{figure*}

In Figure~\ref{fig:qual_examples}, we depicted our model with some stimuli for qualitative evaluation of the model. This experiment and its results illustrate that the model V4 neurons show selectivities to local hues, similar to those patches observed in~\cite{li2014Map}. 

\section{Hue Distance Correlation}
\begin{figure*}[htp]
	\centering
	\subfigure{\includegraphics[width=0.25\textwidth]{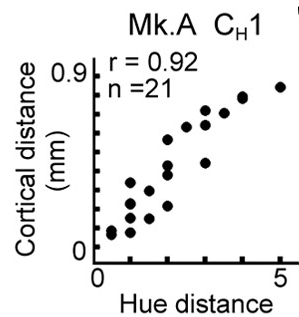}}~
	\subfigure{\includegraphics[width=0.25\textwidth]{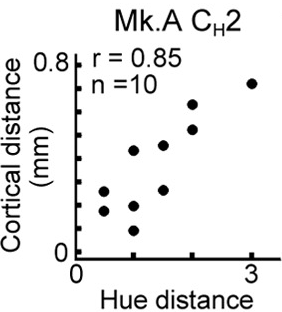}}~
	\subfigure{\includegraphics[width=0.25\textwidth]{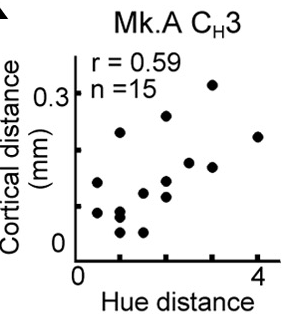}}
	\caption{Cortical distance of activated patches as a function of hue distances in three clusters in V4 (plots adapted from~\protect\cite{li2014Map}).}
	\label{fig:li_hue_distance}
\end{figure*}

In their study, Li~\etal~\cite{li2014Map} found a correlation between the hue distances and the cortical distance of activated patches in each cluster. Figure~\ref{fig:li_hue_distance} depicts three plots from~\cite{li2014Map}, which show the cortical distances of the activated patches as a function of hue distances for three different clusters.
 In these plots, the hue distances vary between 0 and 5 as they first convert each hue to a number in the range $[0, 5]$, with 0 for magenta, 1 for red, 2 for yellow and so on, according to the sequence ordering of patches witnessed in clusters. Then, the hue distances are computed as the difference of these values assigned. In this scheme of representation for hues, magenta and blue are two very distant hues with distance equal to 5, even though these hues are only 60 deg apart on the hue circle as shown in Figure~\ref{fig:hue_wheel}. Moreover, magenta and green hues that are 180 deg apart are at hue distance equal to 3 in this form of representation. In short, their representation of hues does not capture the true distance of hues on the hue circle.

In order to test for a similar relationship between hue distances and the pattern of activities of model V4 neurons, we tested our model with the stimulus shown in Figure~\ref{subfig:ccm_result}. Unlike Li~\etal~\cite{li2014Map}, we represent each hue by its angle on the hue circle, with red starting at 0 deg. Therefore, the longest hue distance is 180 deg. Specifically, yellow and blue that are farthest away from each other on the hue circle have the hue distance of 180 deg. This representation, in contrast with that of Li~\mbox{\cite{li2014Map}}, has the benefit of mapping similar hues to smaller distances. In this case, blue and magenta are only 60 deg apart.

\begin{figure*}[!htp]
	\centering
	\includegraphics[width = 0.48\textwidth,clip]{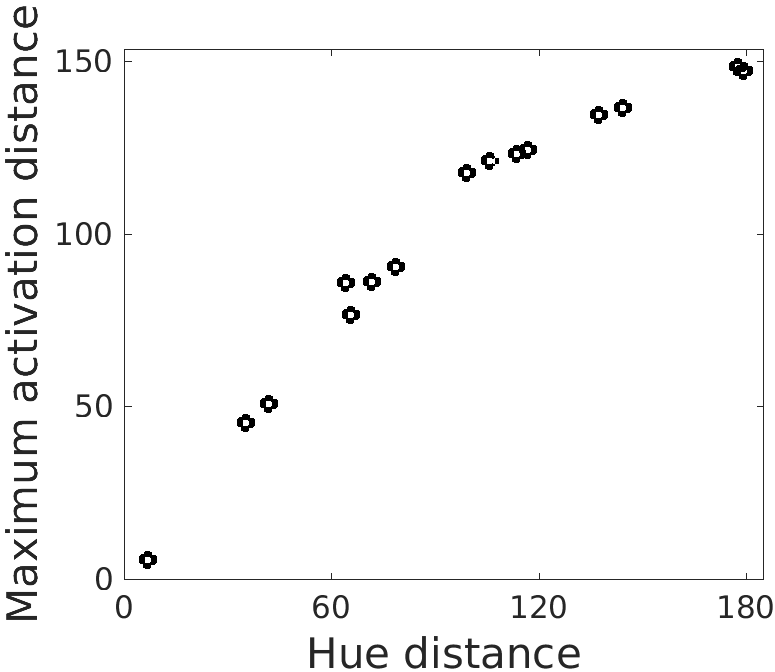}
	\caption{Correlation analysis between the hue distances and maximum activation distances in each hue-selective map in model layer V4. The hue distances are in degrees and the activation distances are in pixels. Each neuron-type map in the model is $256\times256$ pixels.}
	\label{fig:hue_distance}
\end{figure*}

In this experiment, we analyzed the correlation between hue distances on the hue circle and the distance of maximally activated hue-selective neurons in each map. Specifically, we expected to observe that as the hues shift on the presented hue circle stimulus, the maximum activation location in individual model hue-selective neuron maps shifts with a similar pattern and moving from one neuron type to another. Even though this is different from the correlation between hues and cortical patch distances reported in~\cite{li2014Map}, this experiment shows a shift between neurons with peak activities for changes in the hues. For this purpose, we computed six pairs of the form (hue, maximum response location) for each of the six hue-selective neuron maps in model V4 layer. As a result, we had $\binom 62 = 15$ differences computed between each two of such pairs. The plot in Figure~\ref{fig:hue_distance} demonstrates the maximum activation distances as a function of hue distances and exhibits a clear correlation. The correlation coefficient was $r = 0.9582, p = 1.94\times10^{-8}$. 

In conclusion, we observed, in the two aforementioned experiments, that the hue selectivity and pattern of activities of our model V4 neurons resemble those of neurons in V4 of the visual system. Moreover, local hue modeling can be achieved by combining signals from single-opponent cells. In other words, the intermediary hues that model V4 neurons represent are simply obtained from the primary hues that single-opponent cells encode.

\subsection{Hue Reconstruction}
In their work, Li and colleagues~\cite{li2014Map} showed that in monkeys, any hue was represented by 1-4 patches. Moreover, they showed that different hues were encoded with different multi-patch patterns. Then, they suggested that a combination of these activated patches can form a representation for the much larger space of physical colors. 

Along this line, we show, through a few examples, that for a given hue, a linear combination of model V4 neurons can be learned and used for representing that particular hue. It is important to note that it would be impossible to learn weights for the infinitely many possible physical hues. Hence, we show only a few examples here. However, our experiment is an instance of the possible mechanism for color representation suggested by Li~\etal~\cite{li2014Map}.

In this experiment, for a given hue value, we independently sampled the saturation and lightness dimensions at 500 points. The samples were uniformly distributed along each dimension. As a result, we have 500 colors of the same hue. The goal is to compute a linear combination of model V4 neurons, which can reconstruct the groundtruth hue. 

The hues in this experiment were represented as a number in the $(0, 2\pi]$ range. For numerical reasons, red is represented as $2\pi$, not 0. Using the ``stepwiselm" function in MATLAB (MathWorks), we performed a stepwise linear regression on the sampled colors. The choice of stepwise linear regression was made for the following reason. Model V4 neurons are not independent and therefore, the activities of these neurons used for learning are at times highly correlated. For example, magenta is highly correlated with blue with a correlation coefficient of approximately 0.98. This implies that a subset of these neurons is enough to describe the groundtruth hue. Stepwise linear regression serves this purpose. It takes one model V4 neuron at each step and considers adding it to or subtracting it from the model describing the data according to some criterion. A neuron will not be added to the model if it does not significantly improve the error term. 

Table~\ref{table:hue_exp_weights} shows some of the results for this experiment. Interestingly, in all cases, no more than four neuron types were selected, even though no such constraint was imposed. That is, the stepwise linear regression algorithm found that four types of neurons are enough to model the data. This is in agreement with the findings of~\cite{li2014Map}. Moreover, the combination of the selected model V4 neurons spans the RGB space. As an example, in the case of hue at 360 deg, which corresponds to red, in the first step, the model finds yellow neuron responses better describe the data with $p=0$, while red neurons had $p=7.1\times10^{-321}$. Then, red neurons with negative, magenta with positive and green neurons with negative contributions were added sequentially.
\begin{table*}[t]
\small\sf
\centering
\captionsetup{justification=centering}
\caption{The choice of weights for model V4 cells used for hue reconstruction in a few example hues.\label{table:comb_experiment}}
\begin{tabular}{cccccccc}
\hline
Groundtruth hue (deg) &\multicolumn{6}{c}{Model V4 neuron} & RMS error\\
	\cline{2-7}
	& Red & Yellow & Green & Cyan & Blue & Magenta &\\
	\hline
	red (360) &-1.5247&11.6516&-0.8669&0&0&1.8225&$5.89\times 10^{-6}$\\
	\hline
	Yellow (60) &0.9401&-9.7808 &4.2355 & -0.5276 & 0 & 0&$4.82\times 10^{-6}$\\
	\hline
	blue-magenta (270) &0.7823& -1.5262 & 0 & 1.3847 & 0 & -0.3652 & $4.29\times 10^{-6}$\\

\hline
\end{tabular}\\[10pt]
\label{table:hue_exp_weights}
\end{table*}

In the yellow hue example, in the first step of ``stepwiselm'', the model finds both yellow and red neurons equally well describe the data and picks the red neuron in the first step. The next model neuron added is green, which combined with red makes yellow. In the next two steps, yellow and blue model neurons are added to cancel out the imbalance in weights between the initially added red and green neurons.

The last row in Table~\ref{table:hue_exp_weights} is most insightful. It presents the weights for a hue in equal distance from blue (240 deg) and magenta (300 deg). The weights for this example seem counter-intuitive as they include cyan and red with positive contribution and magenta and yellow with negative. In addition, blue is absent. However counterintuitive the weights seem, they start to make sense when one considers the hue circle depicted in Figure~\ref{fig:hue_wheel}. 
The hue at 270 deg, at first, can be seen as one between blue (240 deg) and magenta (300 deg). However, careful scrutiny of the hue circle reveals that this hue at 270 deg can also be considered between cyan (180 deg) and red (360 deg). Hence, this particular hue can be reconstructed by a combination of red and cyan neurons. The imbalance in the weights between red and cyan hues is to some degree canceled by the negative contribution of the magenta and yellow neurons.

Once again, it must be stressed that this experiment was performed to examine the possibility of combinatorial representation mechanisms and a thorough investigation of this mechanism in the computational sense is left for future work. The examples shown here attest to the fact that intermediary hues encoded by model V4 neurons can indeed span the massive space of physical hues and are enough for reconstructing any arbitrary hue from this space.

\section{Discussion}
In this work, we introduced a hierarchical model for local hue representation. This biologically plausible model demonstrates that a network of single-opponent and hue-selective neurons can achieve this purpose. We suggested a mechanistic computational model of neurons that explicitly represent those in LGN, V1, V2, and V4. In other words, not only we have a model for hue-selective neurons, but we have a hierarchy of contributing cells for local hue encoding, each layer individually modeled and analyzed. While our model LGN cells are linear in terms of cone inputs, nonlinearity is increased from one layer to another, starting from V1. Through single-opponent mechanisms, we implemented $\text{L+M-}$, $\text{L-M+}$, $\text{S+(L+M)-}$, and $\text{S-(L+M)+}$ neurons in LGN, V1, and V2. Hue-sensitive neurons in V4 were obtained by input from these four neuron types in V2. 
We presented that the $S+(L+M)-$ and $S-(L+M)+$ single-opponent neurons in our network have wide tuning curves, while $L+M-$ and $L-M+$ cells demonstrate a clear hue sensitivity. In hue-selective layer of V4 neurons, the tuning curves are narrower than those of the previous layers. Similarly, one can model neurons with yet narrower tunings in IT by combining the hue-selective neurons in V4. Moreover, with various weighting combinations, in both V4 and IT, one can obtain a variety of hue-selective cells in these layers. An important implication of these observations is that our model has the capacity of encoding unique hues in V4 or later layers, should we extend our network to beyond V4.

Our experimental results demonstrated that hue selectivity for model V4 neurons similar to that of neurons in layer V4 of the visual system was successfully achieved. In addition, our observations from the hue reconstruction experiment clearly confirmed the possibility of reconstructing the whole hue space using a combination of the hue-selective neurons in the model V4 layer. How this is achieved in the brain, for the infinitely many possible hues, remains to be investigated.

In future, we would like to extend the model to encode saturation and lightness. Moreover, we would like to address the problem of learning weights from V2 to V4. Finally, the experiment on hue reconstruction was performed with a simple linear regression model. A more sophisticated learning algorithm might result in more insightful weights. 

\section*{Acknowledgements}
This research was supported by several sources for which the authors are grateful: Air Force Office of Scientific Research (FA9550-14-1-0393), the Canada Research Chairs Program, and the Natural Sciences and Engineering Research Council of Canada. The authors thank Dr. Rakesh Sengupta for useful discussions during the course of this research.

\bibliographystyle{IEEEtran}
\bibliography{color_references}

\end{document}